\documentclass[10pt,twocolumn,letterpaper]{article}

\usepackage{cvpr}              

\usepackage[dvipsnames]{xcolor}

\usepackage{multirow}
\usepackage{colortbl}
\usepackage{amsthm,amsmath,amssymb}
\usepackage{bbding}
\usepackage{algorithm}
\usepackage{algorithmic}
\usepackage{makecell}

\newcommand{\gc}{\cellcolor[gray]{0.85}}
\newcommand{\rc}{\cellcolor[gray]{0.95}}

\definecolor{cvprblue}{rgb}{0.21,0.49,0.74}
\usepackage[pagebackref,breaklinks,colorlinks,allcolors=cvprblue]{hyperref}

\def\paperID{7259} 
\def\confName{CVPR}
\def\confYear{2025}

\title{Hierarchical Features Matter: A Deep Exploration of Progressive Parameterization Method for Dataset Distillation}

\author{
 Xinhao Zhong$^{1,}$\thanks{Equal Contribution.} \quad Hao Fang$^{3,*}$ \quad Bin Chen$^{1, 2}$\thanks{Corresponding Author.} \quad Xulin Gu$^1$ \quad Meikang Qiu$^4$ \quad Shuhan Qi$^1$ \quad Shu-Tao Xia$^{2, 3}$ \\
$^1$Harbin Institute of Technology, Shenzhen \quad $^2$Peng Cheng Laboratory \\ 
$^3$Tsinghua Shenzhen International Graduate School, Tsinghua University \quad $^4$Augusta University\\
\footnotesize{\texttt{xh021213@gmail.com,}}
    \footnotesize{\texttt{fang-h23@mails.tsinghua.edu.cn,}} 
    \footnotesize{\texttt{chenbin2021@hit.edu.cn, 210110720@stu.hit.edu.cn,}}\\ 
    \footnotesize{\texttt{shuhanqi@cs.hitsz.edu.cn, qiumeikang@yahoo.com, xiast@sz.tsinghua.edu.cn;}}
}

\begin{document}
\maketitle
\begin{abstract}
Dataset distillation is an emerging dataset reduction method, which condenses large-scale datasets while maintaining task accuracy. Current parameterization methods achieve enhanced performance under extremely high compression ratio by optimizing determined synthetic dataset in informative feature domain. However, they limit themselves to a fixed optimization space for distillation, neglecting the diverse guidance across different informative latent spaces. To overcome this limitation, we propose a novel parameterization method dubbed Hierarchical Parameterization Distillation (H-PD), to systematically explore hierarchical feature within provided feature space (e.g., layers within pre-trained generative adversarial networks). We verify the correctness of our insights by applying the hierarchical optimization strategy on GAN-based parameterization method. In addition, we introduce a novel class-relevant feature distance metric to alleviate the computational burden associated with synthetic dataset evaluation, bridging the gap between synthetic and original datasets. Experimental results demonstrate that the proposed H-PD achieves a significant performance improvement under various settings with equivalent time consumption, and even surpasses current generative distillation using diffusion models under extreme compression ratios IPC=1 and IPC=10. Our code is available at \url{https://github.com/ndhg1213/H-PD}
\end{abstract}    
\section{Introduction}
\label{sec:intro}

\begin{figure}[!htbp]
\centering
\resizebox{0.8\linewidth}{!}{
\includegraphics{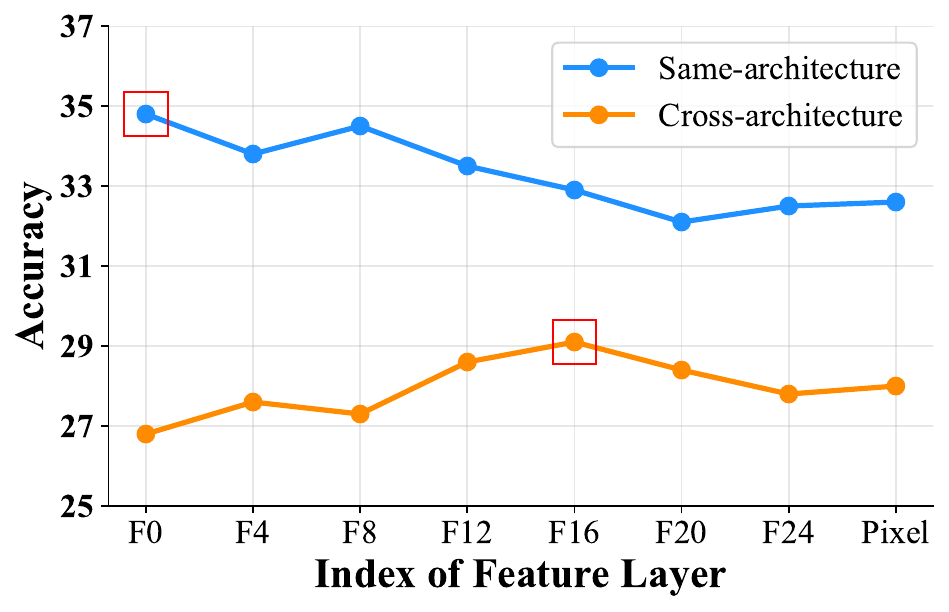}}
    \caption{Performance of synthetic datasets condensed from various feature domains provided by GAN under the same settings (DSA on ImageNet-Birds). }
    \label{not_align}
\vspace{-1.2em}
\end{figure}
In recent years, deep learning has made significant strides in various research fields, encompassing computer vision~\cite{he2016deep,dosovitskiy2020image} and natural language processing~\cite{devlin2018bert,brown2020language}. These advancements have been facilitated by utilizing larger and more intricate deep neural networks (DNNs) in conjunction with numerous datasets tailored for diverse application fields. However, as the complexity of various learning tasks increases, neural networks have grown both deeper and wider, resulting in an exponential surge in the size of datasets required for training these models. This has presented a substantial challenge to data storage and processing efficiency \cite{lei2023comprehensive}, further exacerbating the bottleneck in deep learning due to the mismatch between the enormous data volume and limited computing resources.

Dataset distillation (DD)~\cite{wang2018dataset}  has emerged as a promising solution to the aforementioned issues. It allows for the generation of a more compact synthetic dataset, where each data point encapsulates a higher concentration of task-specific information than its real counterparts. When trained on the synthetic dataset, the network can achieve performance comparable to its counterpart using the original dataset. By significantly reducing the size of the training data, dataset distillation offers a substantial reduction in training costs and memory consumption. Various methods have been proposed to enhance the performance of the condensed dataset. 

Subsequently, synthetic dataset parameterization methods \cite{zhao2022synthesizing,liu2022dataset,cazenavette2023generalizing,shin2023frequency} employ differentiable operations to process synthetic images, shifting the optimization space from pixels to feature domains. These methods benefit from the efficient guidance of hidden features, thus achieving better performance. However, existing parameterization methods focus on one fixed optimization space, overlooking the informative guidance across multiple corresponding feature domains. For example, FreD \cite{shin2023frequency} optimizes the synthetic dataset in the low-frequency space using discrete cosine transform (DCT), while ignoring informative guidance in the high-frequency domain.
In recent years, several recent studies have exploited the rich semantic information encoded in the generators to enhance dataset distillation. With the rapid advancement achieved by generative models, one category of distillation methods utilizes diffusion models~\cite{rombach2022high} to generative informative samples~\cite{gu2024efficient,su2024d}. However, under high compression rates (i.e., IPC=1/10), these methods can degrade into coreset selection methods. Another category of distillation methods employs GANs in an optimization-based manner~\cite{zhao2022synthesizing,cazenavette2023generalizing} to parameterize the synthetic datasets and achieve reliable results. 

In contrast to the aforementioned methods, GAN-based parameterization distillation methods possess an optimization space with richer semantic information. ITGAN\cite{zhao2022synthesizing} directly optimizes the initial latent space of GAN and achieves significant performance improvements on low-resolution datasets. To fully utilize the GAN prior, GLaD~\cite{cazenavette2023generalizing} decomposes the GAN structure and manually selects the intermediate layer, greatly enhancing the cross-architecture performance of the synthetic dataset. However, existing method exhibit a performance decrease in the same-architecture settings when coupled with certain dataset distillation methods as suggested in Figure~\ref{not_align}. In this manner, even though synthetic datasets are condensed from the optimally selected intermediate layer through preliminary experiments by manual picking, the diverse model architectures still lead to changes in the optimal performance. As aforementioned parameterization methods, current GAN-based approaches limit the optimization space to a specific feature domain and necessitate extensive computing time and resources to manually select the optimal feature domain for different settings. 
This naturally raises a question: \emph{\textbf{Does a fixed optimization space meet the demands of dynamic data distribution and model architectures during parameterization dataset distillation?}}

To address this question, we propose a straightforward and efficient approach based on the parameterization prior, Hierarchical Parameterization Distillation (H-PD), which explores the significance of hierarchical features. To verify our intuition, we design a well-design framework on GAN-based parameterization method. The proposed H-PD embraces adaptive exploration across all hierarchical feature domains within GAN models. Specifically, we decompose the GAN structure, undertaking a greedy search that spans different hierarchical feature domains.
During the distillation process, we optimize these hierarchical latents within the GAN model, guided by the loss from the dataset distillation task. Throughout this optimization, we track the best hierarchical latents at the current layer, feeding them into the next layer. This iterative process continues until the optimizer traverses the hierarchical layers and reaches the pixel domain. To mitigate the time-consuming nature of performance evaluation, we introduce a class-relevant feature distance metric between the synthetic and real datasets to search for the optimal latent feature. This metric serves as a performance estimation for the synthetic datasets, encapsulating the significance of hierarchical features. Crucially, our method explores hierarchical features more comprehensively than previous approaches, which only rely on a single fixed feature domain as image priors. 

Our main contributions can be summarized as follows:
\begin{itemize}
    \item We revisited the shortcomings of existing parameterization methods, and provided a novel parametrization framework to enhance their efficiency by leveraging information across various feature domains.
    \item Through a well-designed framework, we effectively enhanced the performance of GAN-based parameterization methods, demonstrating the validity of our insights.
    \item To mitigate the computational demands associated with searching feature domains, we introduce a novel class-relevant feature distance metric, saving valuable computational time by approximating the real performance of the synthetic dataset.
\end{itemize}

\section{Related Work}

\subsection{Dataset Distillation}

Dataset distillation was initially regarded as a meta-learning problem~\cite{wang2018dataset}. It involves minimizing the loss function of the synthetic dataset using a model trained on the synthetic dataset. Since then, several approaches have been proposed to enhance the performance of dataset distillation. One category of methods utilizes ridge regression models to approximate the inner loop optimization~\cite{bohdal2020flexible,nguyen2020dataset,nguyen2021dataset,zhou2022dataset}. Another category of methods selects an informative space as a proxy target to address the unrolled optimization problem. DC~\cite{zhao2020dataset}, DSA~\cite{zhao2021dataset} and DCC~\cite{lee2022dataset} match the weight gradients of models trained on the real and synthetic dataset, while DM~\cite{zhao2023dataset},  CAFE~\cite{wang2022cafe} and DataDAM~\cite{sajedi2023datadam} use feature distance between the real and synthetic dataset as matching targets. MTT~\cite{cazenavette2022dataset} and TESLA~\cite{cui2023scaling} match the model parameter trajectories obtained from training on the real dataset and the synthetic dataset. In recent years, some studies have argued that the bi-level optimization structure required by traditional dataset distillation is redundant and computationally expensive. 

Other studies suggest that the pixel domain, where images reside, is considered a high-dimensional space. Therefore, performance improvement can be achieved by parameterizing the synthetic dataset and transferring the optimization space. IDC~\cite{kim2022dataset} and HaBa~\cite{liu2022dataset} perform optimization in a low-dimensional space using differentiable operations, while GLaD~\cite{cazenavette2023generalizing} and ITGAN~\cite{zhao2022synthesizing} utilize the feature domain provided by GANs as the optimization space, both of them employ pre-trained GANs as priors. FreD~\cite{shin2023frequency} employs traditional compression methods (e.g., DCT) to provide a low-frequency space as the optimization space. However, existing parameterization methods fix the optimization space thus neglecting the guidance from multi-feature domains.

The proposed H-PD introduces an innovative approach to parameterizing synthetic datasets and be verified on GAN-based parameterization methods, representing a broader and more encompassing enhancement compared to previous approaches utilizing fixed optimization space.

\subsection{GAN for parameterization distillation}

GAN~\cite{creswell2018generative} is a deep generative model trained adversarially to learn the distribution of real images. 
Recent studies have shown that GANs can tackle inverse problems by mapping images into their latent space~\cite{xia2022gan,chai2021using, fang2024privacy, fang2023gifd, qiu2024closer, fang2024one}, enabling tasks like image editing~\cite{tewari2020pie,brock2016neural}. 
Incorporating image distribution information into GAN enhances the performance of dataset distillation by utilizing GAN to parameterize the synthetic dataset. 
GLaD employs GAN (e.g., StyleGAN-XL~\cite{sauer2022stylegan}) as a prior and significantly improves the cross-architecture performance of the synthetic dataset by selecting the feature domain of GAN's intermediate layers as the optimization space. However, it overlooks the fact that the optimal optimization space may vary when dealing with different datasets, even with the same dataset distillation method. Additionally, it ignores the guidance offered by GAN's earlier layers. 

Compared with the current dataset distillation methods ~\cite{gu2024efficient, su2024d} based on diffusion models, which could be more suitable for generative methods. GLaD can achieve significantly better performance under extreme compression ratios of IPC=1 and IPC=10 by enhancing the optimization-based methods as an efficient plugin. With the observation, the proposed H-PD further explores hierarchical feature domains of pre-trained GANs to address the limitation of GLaD, resulting in a novel parameterization method that successfully leveraging informative guidance within unfixed optimization space.
\begin{figure*}[!htbp]
\centering
	\subfloat[Fixed optimization space.]{
    \label{fixed space}
    \includegraphics[width = 0.42\textwidth]{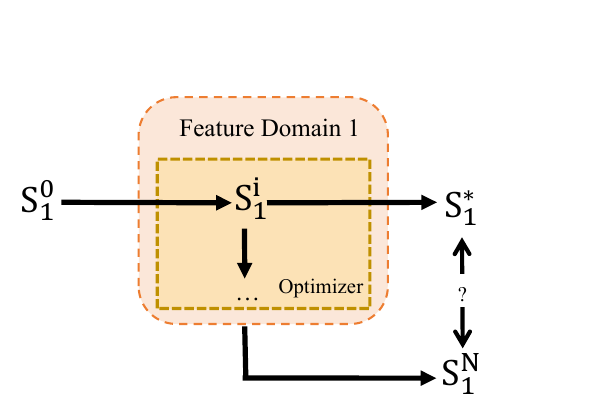}}
	\hfill
	\subfloat[Unfixed optimization space.]{
    \label{unfixed space}
    \includegraphics[width = 0.55\textwidth]{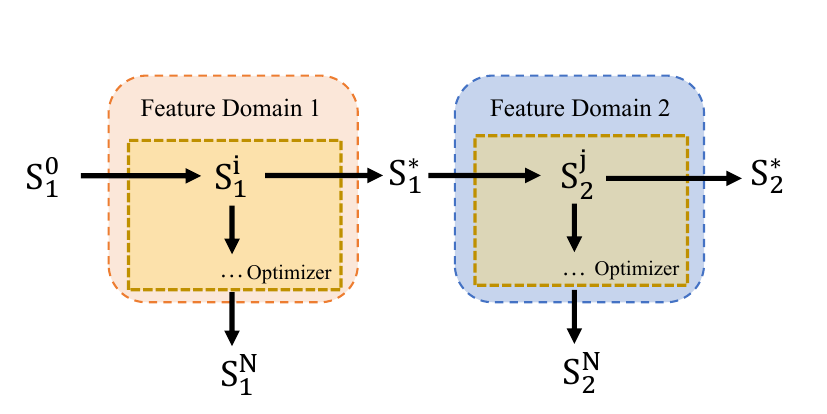}}
\caption{The comparison between fixed optimization space and unfixed optimization space. $\mathcal{S}^{i}$ is the synthetic dataset at optimization steps $i$, $\mathcal{S}^{*}$ is the optimal synthetic dataset selected during the optimization path, $\mathcal{S}_{j}$ is the synthetic dataset optimized in feature domain $j$.}
\label{space comparison}
\vspace{-0.8em}
\end{figure*}
\section{Method}

In this section, we first present the problem definition of dataset distillation and discuss existing methods that parameterize synthetic datasets using GANs. Subsequently, we delve into the specifics of our method, aiming to improve upon previous works by exploring the feature domains provided by GANs. Finally, we propose an alternative evaluation scheme that assesses the synthetic dataset's performance by measuring the layer-wise feature distance between it and the real one. 

\subsection{Preliminaries}

Dataset distillation necessitates a real large-scale dataset $\mathcal{T} = \{ (\mathbf{x}_{t}^{i}, y_{t}^{i}) \}_{i = 1}^{\mathcal{T}}$ and aims to create a smaller synthetic dataset $\mathcal{S} = \{ (\mathbf{x}_{s}^{i}, y_{s}^{i}) \}_{i = 1}^{\mathcal{S}}$ ($\left|\mathcal{S}\right| \ll \left|\mathcal{T}\right|$), minimizing the performance gap between models trained on the two datasets. To achieve this, a well-designed matching objective $\phi(\cdot)$ is employed to extract feature distances in a specific informative space, representing the performance gap between the real and synthetic datasets. The optimization process involves initializing the synthetic dataset from the real dataset and iteratively updating it by minimizing the feature distance, which can be formulated below: 
\begin{align}
    \mathcal{S}^{*} = \mathop{\arg\min}\limits_{\mathcal{S}}\mathcal{M}(\phi(\mathcal{S}), \phi(\mathcal{T})),
\end{align}
where $\mathcal{M}(\,\cdot,\cdot\,)$ denotes some matching metric, e.g., neural network gradients~\cite{zhao2020dataset}, exacted features\cite{zhao2023dataset}, and training trajectories\cite{cazenavette2022dataset}.

Building upon these findings, methods that parameterize synthetic datasets shift the optimization space from the pixel domain to the feature domain by employing differentiable operations. For instance, GAN priors-based methods~\cite{zhao2022synthesizing,cazenavette2023generalizing} can be formulated uniformly below:
\begin{eqnarray}
    \mathbf{z}^{*} = \mathop{\arg\min}\limits_{\mathbf{z}\in\mathcal{Z}}\mathcal{M}(\phi(G_{w}(\mathbf{z})), \phi(\mathcal{T})),
\end{eqnarray}
where $\mathbf{z}\in\mathcal{Z}$ represents the latent in a specific feature domain of a pre-trained generative model $G_{w}(\,\cdot\,)$. Guided by GAN priors, these methods demonstrate substantial performance improvements.

\subsection{Theoretical Insights of Unfixed Optimization Space and Fixed Optimization Space}
\label{vs}
Previous parameterization methods distill knowledge within a fixed optimization space, where the starting point in a given feature domain is typically randomly initialized. In contrast, our proposed H-PD framework performs progressive optimization on the latents. Figure \ref{space comparison} provides a detailed comparison of the optimization processes in fixed versus flexible feature spaces. Beyond the limited guidance across various feature domains, the fixed optimization space encounters a critical bottleneck: it restricts further enhancement by limiting the selection of superior synthetic datasets based on explicit or implicit criteria. In particular, the optimization within a fixed space can be viewed as a continuous process. If a temporarily optimal result, denoted as $S^{i}$, is chosen before convergence, and the corresponding optimization epoch does not mark the end of the optimization trajectory, a fundamental issue arises: \textit{should $S^{i}$ serve as the starting point for the next phase of the optimization, or should the suboptimal result $S^{N}$ be used instead?} Opting for the latter can trap the optimization in a local minimum due to the absence of robust criterion-driven guidance, while choosing the former risks creating an optimization loop, effectively discarding the progress between $S^{i}$ and $S^{N}$ and forcing a reset.

Next, we provide the theoretical insights of its effectiveness. Specifically, we denote the random latent and optimized latent as $\mathbf{z}_\textbf{\textrm{rand}}$ and $\mathbf{z}_\textbf{\textrm{opt}}$, respectively. For an effective dataset distillation method, we assume that the distilled data distribution $P({Z_\textbf{\textrm{lic}}})$ fits the original dataset distribution in a lossless manner. Let the coupling latent $(\mathbf{z}_\textbf{\textrm{rand}}, \mathbf{z}_\textbf{\textrm{opt}})$ be the observed values of random variables $Z_\textbf{\textrm{rand}}$ and $Z_\textbf{\textrm{opt}}$ respectively. Let $c(\cdot)$ be a cost function defined in optimal transport theory~\citep{liu2021lsmi}, which satisfies $\mathbb{E}[c({X} - {Y})] \propto 1 / I(X; Y)$. Then we can obtain:
\begin{eqnarray}  
&&\mathbb{E}[c(Z_\textbf{\textrm{lic}}-Z_\textbf{\textrm{opt}})]\nonumber\\
&=&k/\textrm{D}_\textrm{KL}(P({Z_\textbf{\textrm{lic}}, Z_\textbf{\textrm{opt}}})||P{(Z_\textbf{\textrm{lic}}})\cdot P({Z_\textbf{\textrm{opt}}})) \nonumber\\
&=&k/[H(Z_\textbf{\textrm{lic}})-H(Z_\textbf{\textrm{lic}}|Z_\textbf{\textrm{opt}})]\nonumber\\
\label{eq: leq}
&\leq& k/[H(Z_\textbf{\textrm{lic}})-H(Z_\textbf{\textrm{lic}}|Z_\textbf{\textrm{rand}})] \\
&=& k/\textrm{D}_\textrm{KL}(P({Z_\textbf{\textrm{lic}}, Z_\textbf{\textrm{rand}}})||P{(Z_\textbf{\textrm{lic}}})\cdot P({Z_\textbf{\textrm{rand}}}))  \nonumber\\
&=&\mathbb{E}[c(Z_\textbf{\textrm{lic}}-Z_\textbf{\textrm{rand}})],\nonumber
\end{eqnarray}
where $k$ denotes a constant, $D_\textrm{KL}(\cdot||\cdot)$ and $H(\cdot)$ stand for Kullback-Leibler divergence and entropy, respectively. The Inequality \eqref{eq: leq} follows since the conditional entropy $H(Z_\textbf{\textrm{lic}}|Z_\textbf{\textrm{opt}})$ is smaller than $H(Z_\textbf{\textrm{lic}}|Z_\textbf{\textrm{rand}})$.

The theoretical analysis indicates that the proposed H-PD method of selecting partially optimized latents, rather than random initialization, reduces optimization cost across different spaces. By leveraging multiple feature domains, it accelerates convergence and, with its implicit evaluation criterion, effectively avoids local optima—an issue common in fixed optimization spaces.

\subsection{Progressive Optimization with Hierarchical Feature Domains}
\label{hierarchical feature}
To verify our insights about the utilization of unfixed parameterization optimization space, we apply it with GAN-based parameterization method (i.e., GLaD~\cite{cazenavette2023generalizing}).
As depicted in Algorithm~\ref{alg:algorithm1}, our approach diverges from restricting the optimization space to a specific feature domain of the GANs. Instead, we aim to explore the hierarchical layers of the GAN, striving to enhance the effective utilization of the prior information.

To sufficiently utilize the informative guidance from the hierarchical feature domains, we decompose a pre-trained GAN $G_{w}(\cdot)$ for hierarchical layer searching, i.e.,
\begin{eqnarray}
G_{w}(\,\cdot\,) = G_{K-1}\circ G_{K-2}\circ\cdots\circ G_{1}\circ G_{0}(\,\cdot\,).
\end{eqnarray}
For each hierarchical layer $G_{i}$ provided by GAN, we repeat the following steps. Firstly, we generate images from $\mathbf{z}_{i}$ only using the remaining synthesizing network $G_{k-1}(\,\cdot\,)\circ G_{k-2}(\,\cdot\,)\circ\cdots\circ G_{i}(\,\cdot\,)$. Then, we employ the distillation method (e.g., MTT~\cite{cazenavette2022dataset}) to calculate $\mathcal{L}$ based on the synthetic dataset composed of generated images and the real dataset to optimize $\mathbf{z}_{i}$ with an SGD optimizer, the optimization process lasts for a pre-determined and fixed $N$ steps. After completing the optimization process for a specific layer, we implicitly evaluate the latents synthesized during the SGD optimization process and record $\mathbf{z}_{i}^{*}$ as the optimal latents for the current layer.  Finally, we pass $\mathbf{z}_{i}^{*}$ into the next layer to obtain $\mathbf{z}_{i+1}^{0}$ as the initial latent for the next layer.

When the optimization space reaches the ultimate pixel domain, we choose the optimal latent $\mathbf{z}^{*}$ from the recorded latents $\mathbf{z}_{i}^{*}$ based on the real performance of the synthetic dataset $\mathcal{S}$ generated by the corresponding remaining synthesizing network $G_{k-1}(\,\cdot\,)\circ G_{k-2}(\,\cdot\,)\circ\cdots\circ G_{i}(\,\cdot\,)$. In this way, we fully explore the feature domain of the GAN, leveraging its rich information. 

\begin{algorithm}[t]
    \caption{Pseudocode of H-PD with GAN-based parameterization method}
    \label{alg:algorithm}
    \textbf{Input}: $G_{w}(\,\cdot\,)$: a pre-trained generative model; $K$: the number of hierarchical layers; $N$: distillation steps; $\mathcal{T}$: real training dataset; $P_{z}$: distribution of latent initializations; $\mathcal{L}$: distillation loss; $Acc(\,\cdot\,)$: evaluate real performance of synthetic dataset;
    \begin{algorithmic}[1] 
        \STATE Initial average latent $\mathbf{z} \sim P_{\mathcal{Z}}$
        \STATE Dissemble $G_{w}(\,\cdot\,)$ into $G_{k-1}\circ G_{k-2}\circ\cdots\circ G_{0}(\,\cdot\,)$
        \STATE $acc_{max} = 0$                      
        \FOR {i ${\leftarrow}$ 0 to $K - 1$}
        \STATE $d_{min} = \mathcal{D}(G_{k-1}\circ \cdots\circ G_{i}(\mathbf{z}_{i}^{0}), \mathcal{T})$
        \FOR {j ${\leftarrow}$ 0 to $N - 1$}
        \STATE $\mathcal{S}_{i}^{j} = G_{k-1}\circ \cdots\circ G_{i}(\mathbf{z}_{i}^{j})$.
        \STATE $\mathcal{L} = \mathcal{M}(\phi(\mathcal{S}_{i}^{j}), \phi(\mathcal{T}))$
        \STATE $\mathbf{z}_{i}^{j + 1} \leftarrow SGD(\mathbf{z}_{i}^{j}; \mathcal{L})$
        \IF {$\mathcal{D}(\mathcal{S}_{i}^{j}, \mathcal{T}) \leq d_{min}$}
        \STATE $\mathbf{z}_{i}^{*} = \mathbf{z}_{i}^{j}, \mathcal{S}_{i}^{*} = \mathcal{S}_{i}^{j}$
        \ENDIF
        \ENDFOR
        \IF {$Acc(\mathcal{S}_{i}^{*}) > acc_{max}$}
        \STATE $acc_{max} = Acc(\mathcal{S}_{i}^{*})$
        \STATE $\mathcal{S} = \mathcal{S}_{i}^{*}$
        \ENDIF
        \STATE $\mathbf{z}_{i + 1}^{0} = G_{i}(\mathbf{z}_{i}^{*})$
        \ENDFOR\\
    \end{algorithmic}
    \textbf{Output}: Synthetic dataset $\mathcal{S}$
    \label{alg:algorithm1}
\end{algorithm}
\subsection{Enhancing Performance with Efficient Searching Strategy}

\paragraph{Ensemble-Averaging Latent Initialization}
\label{average initialization}

To mitigate the undesirable time overhead brought by existing methods~\cite{liu2023dream} using clustering or GAN inversion~\cite{xia2022gan}, we propose an inactive searching initialization by calculating the average value of multiple noises, and passing it through the GAN's mapping network to obtain the initial latent $\mathbf{z}_{0}$ with reduced bias. our method showcases simplicity without compromising effectiveness, as confirmed by sufficient experimental results.
\vspace{-0.4em}
\paragraph{Class-relevant Feature Distance}
\label{feature distance}

To search for optimal latent as the optimization starting point of the subsequent feature domain, an efficient implicit evaluation metric is needed to replace the time-consuming evaluation of the synthetic dataset's real performance. We first attempt to use the loss value as a substitute evaluation metric. However, it fails to yield desired results and, in some cases, performs even worse than not searching at all.

To utilize gradient information while maintaining diversity, we adopt the class activation map (CAM) \cite{muhammad2020eigen} by utilizing the gradients of the corresponding class with respect to the feature maps to localize the class-specific features. With the output logits $q = {f}^{d}({w}^{d}; \mathbf{z})$ from the classifier ${w}^{d}$, the CAM is defined as the gradients of output logits ${q}^{y}$ of class $y$ with respect to features $\mathbf{z}$ as follows:
\begin{align}
{g}_\mathbf{z}=\frac{\partial {q}^{y}}{\partial \mathbf{z}}.
\end{align}
To focus attention on the class-relevant region, we propose a novel class-relevant feature distance $\mathcal{D}(\mathcal{S},\mathcal{T})$ between the real dataset $\mathcal{S}$ and the synthetic dataset $\mathcal{T}$. i.e., 
\begin{align}
\nonumber \mathcal{D}(\mathcal{S},\mathcal{T})= & \|\frac{1}{|\mathcal{T}|} \sum_{i=1}^{|\mathcal{T}|} {w}^{e}\left(\mathbf{x}_{t}^{i}\right) \cdot \operatorname{ReLU}({g}_\mathbf{z}^{t}) \\
-  & \frac{1}{|\mathcal{S}|} \sum_{j=1}^{|\mathcal{S}|} {w}^{e}\left(\mathbf{x}_{s}^{j}\right) \cdot \operatorname{ReLU}({g}_\mathbf{z}^{s}) \|^{2},
\end{align}
where ${w}^{e}(\cdot)$ represents the feature extractor of a pre-trained network, and $\operatorname{ReLU}(\cdot)$ is the rectified linear unit function.
\begin{table*}[!htbp]
    \centering
    \resizebox{\textwidth}{!}{
    \begin{tabular}{cccccccccccc}
        \toprule
        Alg. & Method & ImNet-A & ImNet-B & ImNet-C & ImNet-D & ImNet-E & ImNette & ImWoof & ImNet-Birds & ImNet-Fruits & ImNet-Cats\\
        \midrule
            & Pixel  & 51.7\small{$\pm0.2$}     & 53.3\small{$\pm1.0$}     & 48.0\small{$\pm0.7$}     &43.0\small{$\pm0.6$}      & 39.5\small{$\pm0.9$}     & 41.8\small{$\pm0.6$}     & 22.6\small{$\pm0.6$}    & 37.3\small{$\pm0.8$}       & 22.4\small{$\pm1.1$}       & 22.6\small{$\pm0.4$}\\
        TESLA & GLaD   & 50.7\small{$\pm0.4$}     & 51.9\small{$\pm1.3$}     & 44.9\small{$\pm0.4$}     & 39.9\small{$\pm1.7$}     & 37.6\small{$\pm0.7$}     & 38.7\small{$\pm1.6$}     & 23.4\small{$\pm1.1$}    & 35.8\small{$\pm1.4$}       & 23.1\small{$\pm0.4$}       & 26.0\small{$\pm1.1$}\\
            & \gc H-PD   &\gc \textbf{55.1}\small{$\pm0.6$}     &\gc \textbf{57.4}\small{$\pm0.3$}      &\gc \textbf{49.5}\small{$\pm0.6$}     &\gc \textbf{46.3}\small{$\pm0.9$}       &\gc \textbf{43.0}\small{$\pm0.6$}     &\gc \textbf{45.4}\small{$\pm1.1$}     &\gc \textbf{28.3}\small{$\pm0.2$}      &\gc \textbf{39.7}\small{$\pm0.8$}          &\gc \textbf{25.6}\small{$\pm0.7$}          &\gc \textbf{29.6}\small{$\pm1.0$}\\
        \midrule
            & Pixel  & 43.2\small{$\pm0.6$}   & 47.2\small{$\pm0.7$}     & 41.3\small{$\pm0.7$}     & 34.3\small{$\pm1.5$}       & 34.9\small{$\pm1.5$}     & 34.2\small{$\pm1.7$}     & 22.5\small{$\pm1.0$}     & 32.0\small{$\pm1.5$}    & 21.0\small{$\pm0.9$}      & 22.0\small{$\pm0.6$}  \\
        DSA  & GLaD   & 44.1\small{$\pm2.4$}     & 49.2\small{$\pm1.1$}     & 42.0\small{$\pm0.6$}       & 35.6\small{$\pm0.9$}     & 35.8\small{$\pm0.9$}     & 35.4\small{$\pm1.2$}     & 22.3\small{$\pm1.1$}    & 33.8\small{$\pm0.9$}         & 20.7\small{$\pm1.1$}          & 22.6\small{$\pm0.8$}\\
            &\gc H-PD   &\gc \textbf{46.9}\small{$\pm0.8$}      &\gc \textbf{50.7}\small{$\pm0.9$}     & \gc \textbf{43.9}\small{$\pm0.7$}     & \gc \textbf{37.4}\small{$\pm0.4$}     & \gc \textbf{37.2}\small{$\pm0.3$}     & \gc
            \textbf{36.9}\small{$\pm0.8$}     &\gc \textbf{24.0}\small{$\pm0.8$}    &\gc \textbf{35.3}\small{$\pm1.0$}           &\gc \textbf{22.4}\small{$\pm1.1$}          &\gc \textbf{24.1}\small{$\pm0.9$}\\
        \midrule
            & Pixel  & 39.4\small{$\pm1.8$}       & 40.9\small{$\pm1.7$}     & 39.0\small{$\pm1.3$}     & 30.8\small{$\pm0.9$}     & 27.0\small{$\pm0.8$}     & 30.4\small{$\pm2.7$}     & 20.7\small{$\pm1.0$}    & 26.6\small{$\pm2.6$}        & 20.4\small{$\pm1.9$}        & 20.1\small{$\pm1.2$}\\
        DM  & GLaD   & 41.0\small{$\pm1.5$}       & 42.9\small{$\pm1.9$}     & 39.4\small{$\pm1.7$}     & 33.2\small{$\pm1.4$}     & 30.3\small{$\pm1.3$}     & 32.2\small{$\pm1.7$}     & 21.2\small{$\pm1.5$}    & 27.6\small{$\pm1.9$}         & 21.8\small{$\pm1.8$}          & 22.3\small{$\pm1.6$}\\
            &\gc H-PD   &\gc \textbf{42.8}\small{$\pm1.2$}     &\gc \textbf{44.7}\small{$\pm1.3$}     &\gc \textbf{41.1}\small{$\pm1.3$}     & \gc\textbf{34.8}\small{$\pm1.5$}     &\gc \textbf{31.9}\small{$\pm0.9$}     &\gc \textbf{34.8}\small{$\pm1.0$}     &\gc \textbf{23.9}\small{$\pm1.9$}    &\gc \textbf{29.5}\small{$\pm1.5$}         &\gc \textbf{24.4}\small{$\pm2.1$}          &\gc \textbf{24.2}\small{$\pm1.1$}\\
        \bottomrule
    \end{tabular}}
    \caption{Synthetic dataset same-architecture performance (\%) on ImageNet-Subset ($128 \times 128$) under IPC=1. "Pixel"refers to not deploying GAN.}
    \vspace{-1.2em}
    \label{same-architecture performance}
\end{table*}

\begin{table}[!htbp]
    \centering
    \resizebox{\linewidth}{!}{%
     \begin{tabular}{cccc|cc}
        \toprule
        \multirow{2}{*}{Alg.} & \multirow{2}{*}{Method} &  \multicolumn{2}{c|}{Tiny-ImageNet} & \multicolumn{2}{c}{ImageNet-1K} \\
        \cmidrule(lr){3-4} \cmidrule(lr){5-6}
         & & IPC-1 & IPC-10  & IPC-1 & IPC-10\\
        \midrule
             &Pixel & 2.6\small{$\pm0.1$}      & 16.1\small{$\pm0.2$}     & 0.1\small{$\pm0.1$}     & 21.3\small{$\pm0.6$}   \\
         SRe$^2$L &GLaD  & 3.1\small{$\pm0.3$} & 15.7\small{$\pm0.2$} & 1.2\small{$\pm0.1$} & 21.9\small{$\pm0.8$}\\
         &\gc  H-PD  &\gc  \textbf{4.5}\small{$\pm1.0$} &\gc  \textbf{18.3}\small{$\pm0.5$} &\gc  \textbf{2.6}\small{$\pm0.2$} &\gc  \textbf{23.5}\small{$\pm0.4$} \\
        \bottomrule
    \end{tabular}}
    \caption{Synthetic dataset performance comparison with SRe$^2$L on Tiny-ImageNet (64x64) and ImageNet-1K (224x224) evaluted by ResNet-18.} 
    \label{more}
    \vspace{-0.8em}
\end{table}
\section{Experiments}
To verify the efficiency of our proposed method, we conduct experiments using code derived from the open-source GLaD\footnote{ https://georgecazenavette.github.io/glad}. We utilize ImageNet-1K~\cite{deng2009imagenet} subsets and CIFAR-10~\cite{krizhevsky2009learning} to generate high-resolution and low-resolution distilled datasets respectively, with StyleGAN-XL as the deep generative network. To ensure a fair comparison, we maintain consistency by adopting the same network architecture and employing identical hyperparameters. Our code is availabel at https://github.com/ndhg1213/H-PD.

\subsection{Settings and Implementation Details}

\paragraph{Datasets and Network Architectures}
In this study, we build upon previous research by utilizing CIFAR10 and Tiny-ImageNet~\cite{le2015tiny} as low-resolution dataset, For high-resolution experiments, we choose ImageNet-1K and ten subsets from it. These subsets, each consisting of ten categories, are divided into the training and validation sets. 
The detailed categories combination can be found in Appendix.

For the surrogate model for dataset distillation, we choose ConvNet-5~\cite{gidaris2018dynamic} as the backbone network for DM, DSA and TESLA. 
For ImageNet-1K and Tiny-Imagenet, we conduct experiments on SRe$^2$L~\cite{yin2024squeeze} and adopt ResNet-18~\cite{he2016deep} as backbone. To evaluate the performance of the synthetic dataset, we employ various models, including ConveNet, AlexNet~\cite{krizhevsky2012imagenet}, VGG-11~\cite{simonyan2014very}, ResNet-18, and a Vision Transformer model~\cite{dosovitskiy2020image} from the DC-BENCH~\cite{cui2022dc} resource. It is important to note that all of these evaluation models are versions specifically tailored for corresponding resolution datasets.

\subsection{Performance Improvements}
\begin{figure*}[!htbp]
    \centering
    \resizebox{\textwidth}{!}{
    \includegraphics{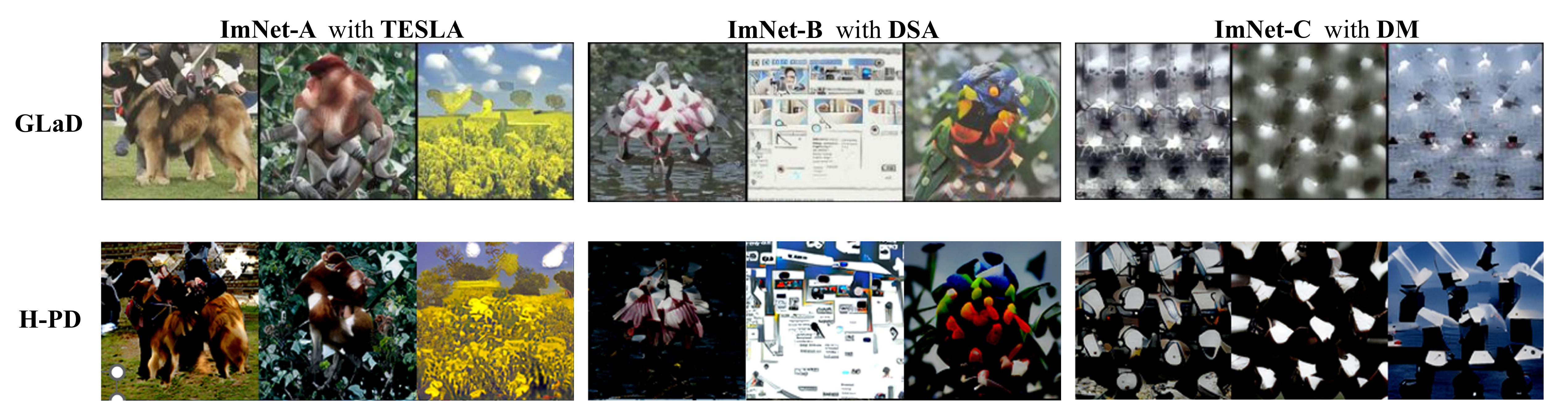}}
    \caption{Visualization comparison of the synthetic datasets with different distillation methods.}
    \label{visualization}
\end{figure*}

The performance comparison of our method with previous works GLaD is shown in Table \ref{same-architecture performance} and Table \ref{more}. We report the same-architecture performance of the synthetic dataset. Since GLaD did not conduct experiments on SRe$^2$L, we chose to adopt the same layers settings (i.e., 12) as TESLA to ensure fairness. compared to optimizing only in a fixed feature space, it can be observed that our method achieves consistent and significant improvements with all the optimization-based methods. This indicates that our method successfully leverages the guidance information provided by all feature domains. The visualized images of synthetic datasets are depicted in Figure \ref{visualization}. 

\begin{table}[!htbp]
    \centering
    \resizebox{\linewidth}{!}{%
     \begin{tabular}{ccccccc}
        \toprule
        Alg. & Method  & ImNet-A  & ImNet-B & ImNet-C & ImNet-D & ImNet-E \\
        \midrule
             &Pixel & 52.3\small{$\pm0.7$}      & 45.1\small{$\pm8.3$}     & 40.1\small{$\pm7.6$}     & 36.1\small{$\pm0.4$}    & 38.1\small{$\pm0.4$}    \\
         DSA &GLaD  & 53.1\small{$\pm1.4$} & 50.1\small{$\pm0.6$} & 48.9\small{$\pm1.1$} & 38.9\small{$\pm1.0$} & 38.4\small{$\pm0.7$}\\
         &\gc  H-PD  &\gc  \textbf{54.1}\small{$\pm1.2$} &\gc  \textbf{52.0}\small{$\pm1.1$} &\gc  \textbf{49.5}\small{$\pm0.8$} &\gc  \textbf{39.8}\small{$\pm0.7$} & \gc \textbf{40.1}\small{$\pm0.7$}\\
         \midrule
                    &Pixel &  52.6\small{$\pm0.4$}     & 50.6\small{$\pm0.5$}     &  47.5\small{$\pm0.7$}    & 35.4\small{$\pm0.4$}    & 36.0\small{$\pm0.5$}    \\
        DM &GLaD  & 52.8\small{$\pm1.0$} & 51.3\small{$\pm0.6$} & 49.7\small{$\pm0.4$} & 36.4\small{$\pm0.4$} & 38.6\small{$\pm0.7$}\\
         &\gc  H-PD  &\gc  \textbf{55.1}\small{$\pm0.5$} &\gc  \textbf{54.2}\small{$\pm0.5$} &\gc  \textbf{50.8}\small{$\pm0.4$} &\gc  \textbf{37.6}\small{$\pm0.6$} &\gc  \textbf{39.9}\small{$\pm0.7$}\\
        \bottomrule
    \end{tabular}}
    \caption{Synthetic dataset cross-architecture performance (\%) on ImageNet-Subset under IPC=10.} 
    \label{ipc10 performance}
    \vspace{-0.8em}
\end{table}

\subsection{More Comparisons with GLaD}
To align the proposed H-PD with GLaD on higher IPC, i.e. only $\text{IPC}=10$ under DSA and DM is reported from original paper, our current confirmatory trials achieves a performance improvement of 1\% to 3\% compared to GLaD with $\text{IPC}=10$ under DSA and DM as shown in Table \ref{ipc10 performance}, respectively, which demonstrates the effectiveness of H-PD.

\begin{table}[!htbp]
    \centering
    \resizebox{\linewidth}{!}{%
      \begin{tabular}{ccccccc}
        \toprule
        Alg.  & Method & ConvNet  & AlexNet  & ResNet-18  & VGG-11  & ViT\\
        \midrule
        &  Pixel &\gc \textbf{46.3}\small{$\pm0.8$} & 26.8\small{$\pm0.6$} & 23.4\small{$\pm1.3$} & 24.9\small{$\pm0.8$} & 21.2\small{$\pm0.4$}\\
   TESLA     &  GLaD &35.5\small{$\pm0.6$} & 27.9\small{$\pm0.6$} & 30.2\small{$\pm0.6$} & 31.3\small{$\pm0.7$} & 22.7\small{$\pm0.4$}\\
        &\gc  H-PD   & 37.2\small{$\pm0.4$}     &\gc  \textbf{28.5}\small{$\pm0.3$}      &\gc  \textbf{31.4}\small{$\pm0.4$}     &\gc  \textbf{32.2}\small{$\pm0.2$}       &\gc  \textbf{24.1}\small{$\pm0.4$}\\
        \midrule             
        &  Pixel  & 28.3\small{$\pm0.3$} & 25.9\small{$\pm0.2$} & 27.3\small{$\pm0.5$} & 28.0\small{$\pm0.5$} & 22.9\small{$\pm0.3$}\\
    DSA     & GLaD  & 29.2\small{$\pm0.8$} & 26.0\small{$\pm0.7$} & 27.6\small{$\pm0.6$} & \gc \textbf{28.2}\small{$\pm0.6$} & 23.4\small{$\pm0.2$}\\
         &\gc  H-PD  &\gc  \textbf{30.2}\small{$\pm0.5$}      &\gc  \textbf{26.6}\small{$\pm0.4$}     &\gc  \textbf{28.2}\small{$\pm0.4$}     & 28.0\small{$\pm0.6$}     & \gc \textbf{24.4}\small{$\pm0.5$}\\        
        \midrule        
        & Pixel   & 26.0\small{$\pm0.6$} & 22.9\small{$\pm0.2$} & 22.2\small{$\pm0.7$} & 23.8\small{$\pm0.5$} & 21.3\small{$\pm0.5$}\\
    DM         & GLaD  & 27.1\small{$\pm0.7$} & 25.1\small{$\pm0.5$} & 22.5\small{$\pm0.7$} & 24.8\small{$\pm0.8$} & 23.0\small{$\pm0.1$} \\
         &\gc  H-PD   &\gc  \textbf{27.6}\small{$\pm0.7$}     &\gc  \textbf{27.5}\small{$\pm0.6$}     &\gc  \textbf{25.6}\small{$\pm0.6$}     & \gc \textbf{25.4}\small{$\pm0.8$}     & \gc \textbf{23.6}\small{$\pm0.5$}\\     \bottomrule
        \end{tabular}}
    \caption{Performance(\%) across different models on CIFAR-10.} 
    \label{CIFAR-10 performance}
    \vspace{-1.2em}
\end{table}

The cross-architecture performance on CIFAR-10~\cite{krizhevsky2009learning} is shown in Table \ref{CIFAR-10 performance}. The results demonstrate that using a shallower StyleGAN-XL structure on the lower-resolution dataset CIFAR-10, H-PD still improves the performance of synthetic datasets distilled by different distillation methods. Please note that the released code of GLaD does not include the data augmentation and hyperparameter settings used by TESLA on CIFAR10, which leads to a poor performance on ConvNet.

To present a more comprehensive comparison, we assess the performance of the synthetic dataset across different architectures as shown in Table \ref{cross-architecture performance}. The cross-architecture accuracy is calculated by averaging the performance of the remaining four models, excluding the backbone model. The results of previous studies are acquired directly from the original papers.
\begin{table}[!htbp]
    \centering
    \resizebox{\linewidth}{!}{
    \begin{tabular}{ccccccc}
        \toprule
        Alg.  & Method & ImNet-A  & ImNet-B  & ImNet-C  & ImNet-D  & ImNet-E \\
        \midrule
            & Pixel   & 33.4\small{$\pm1.5$}     & 34.0\small{$\pm3.4$}     & 31.4\small{$\pm3.4$}     & 27.7\small{$\pm2.7$}     & 24.9\small{$\pm1.8$} \\
        TESLA & GLaD   & 39.9\small{$\pm1.2$}     & 39.4\small{$\pm1.3$}     & 34.9\small{$\pm1.1$}     & 30.4\small{$\pm1.5$}     & 29.0\small{$\pm1.1$} \\
            &\gc H-PD   & \gc \textbf{40.2}\small{$\pm0.3$}     & \gc \textbf{39.8}\small{$\pm0.8$}      & \gc \textbf{35.8}\small{$\pm0.7$}     & \gc \textbf{31.2}\small{$\pm1.0$}       & \gc \textbf{29.5}\small{$\pm0.7$}  \\
        \midrule
            & Pixel   & 38.7\small{$\pm4.2$}     & 38.7\small{$\pm1.0$}     & 33.3\small{$\pm1.9$}     & 26.4\small{$\pm1.1$}     & 27.4\small{$\pm0.9$}\\
        DSA  & GLaD   & 41.8\small{$\pm1.7$}     & 42.1\small{$\pm1.2$}     & 35.8\small{$\pm1.4$}       & 28.0\small{$\pm0.8$}     & 29.3\small{$\pm1.3$} \\
            &\gc  H-PD   &\gc  \textbf{42.4}\small{$\pm1.2$}      & \gc \textbf{42.6}\small{$\pm1.1$}     & \gc \textbf{36.1}\small{$\pm1.1$}     & \gc \textbf{28.7}\small{$\pm1.1$}     & \gc \textbf{29.6}\small{$\pm0.9$} \\
        \midrule
            & Pixel   & 27.2\small{$\pm1.2$}     & 24.4\small{$\pm1.1$}     & 23.0\small{$\pm1.4$}     & 18.4\small{$\pm0.7$}     & 17.7\small{$\pm0.9$}\\
        DM  & GLaD   & 31.6\small{$\pm1.4$}       & 31.3\small{$\pm3.9$}     & 26.9\small{$\pm1.2$}     & 21.5\small{$\pm1.0$}     & 20.4\small{$\pm0.8$}  \\
            &\gc  H-PD   & \gc \textbf{34.9}\small{$\pm2.1$}     & \gc \textbf{33.8}\small{$\pm2.0$}     & \gc \textbf{27.8}\small{$\pm1.7$}     & \gc \textbf{23.6}\small{$\pm1.5$}     & \gc \textbf{22.5}\small{$\pm1.3$} \\
        \bottomrule
    \end{tabular}}
    \caption{Synthetic dataset cross-architecture performance (\%) on ImageNet-Subset.}
    \label{cross-architecture performance}
    \vspace{-0.8em}
\end{table}
\subsection{Comparison with Diffusion Model Based Methods}

Other generative distillation methods consider diffusion as an image generator rather than parameterization methods. Minimax~\cite{gu2024efficient} and D$^4$M~\cite{su2024d} fine-tune the diffusion model and corresponding latent respectively to generate entirely new samples. However, these methods only achieve random sampling outcomes under a high compression ratio (e.g., IPC=1/10), Contrary to these diffusion-based methods, our method in conjunction with GLAD belongs to an optimization-based method, which can effectively accomplish distillation in a more efficient way.

For a fair comparison, we adhere to GLaD's experimental setting (e.g., a 128x128 resolution image corresponds to a one-hot label), disregarding training time matching~\cite{shen2022fast} and mixiup~\cite{zhang2017mixup} strategy. As shown in Table \ref{compare-diff}, our method demonstrates significant advantages at a mechanism compression ratio of IPC=1 without requiring fine-tuning of the pre-trained model.
\begin{table}[!htbp]
    \centering
    \resizebox{\linewidth}{!}{%
     \begin{tabular}{cccccc}
        \toprule
        Method  & ImNette  & ImWoof & ImNet-Birds & ImNet-Fruits & ImNet-Cats \\
        \midrule
            Minimax  & 22.8\small{$\pm0.5$} & 17.8\small{$\pm0.3$} & 23.2\small{$\pm0.7$} & 17.5\small{$\pm0.5$} & 19.8\small{$\pm0.3$}\\
             D$^4$M  & 15.2\small{$\pm0.5$} & 17.4\small{$\pm0.6$} & 18.2\small{$\pm0.4$} & 17.6\small{$\pm0.6$} & 23.4\small{$\pm0.7$}\\
         \gc  H-PD  &\gc  \textbf{45.4}\small{$\pm1.1$} &\gc  \textbf{28.3}\small{$\pm0.2$} &\gc  \textbf{39.7}\small{$\pm0.8$} &\gc  \textbf{25.6}\small{$\pm0.7$} & \gc \textbf{29.6}\small{$\pm1.0$}\\
        \bottomrule
    \end{tabular}}
    \caption{Comparison with generative dataset distillation methods on ImageNet-Subsets under IPC=1.} 
    \label{compare-diff}
    \vspace{-0.8em}
\end{table}
\subsection{Evaluation Protocol}
\label{evaluation protocol}
The approach to assessing the performance of a synthetic dataset is as follows: firstly, a set of models is trained using the synthetic dataset. Once the training is complete, the trained models are validated using the corresponding validation set from the real dataset. For a specific model architecture, this process is repeated five times, and the average performance is calculated based on these repetitions.
\begin{table}[!htbp]
\centering
\tabcolsep=0.5cm
    \resizebox{\linewidth}{!}{
    \begin{tabular}{ccccc}
        \toprule
        Metric & Method  & TESLA  & DSA & DM \\
        \midrule
        \multirow{2}{*}{Time}
        &GLaD  & 75 & \gc \textbf{69} & 64\\
        &\gc H-PD  & \gc \textbf{70} & 73 & \gc \textbf{15}\\
        \midrule
        \multirow{2}{*}{perfomance}
        &GLaD & 45.0{$\pm0.9$} & 41.3{$\pm1.2$}& 37.4{$\pm1.6$}\\
        &\gc H-PD&\gc \textbf{50.3}{$\pm0.6$}&\gc  \textbf{43.2}{$\pm0.6$}&\gc  \textbf{39.1}{$\pm1.2$}\\
        \bottomrule
\end{tabular}}
    \caption{Time complexity (min) and performance (\%) averaged on ImageNet-[A, B, C, D, E].} 
    \label{complexity}
    \vspace{-0.8em}
\end{table}
In previous studies, the evaluation method involved continuously optimizing the entire distillation process for 1000 epochs, with sampling the synthetic dataset every 100 epochs. The best performance among all sampled examples would then be selected. To ensure a fair comparison, we decompose StyleGAN-XL into $G_{11}\circ\cdots\circ G_{1}\circ G_{0}(\,\cdot\,)$ and apply the same optimizer and learning rate for each layer, optimizing for 100 steps. This ensures that the total number of optimization epochs remains consistent, thereby preventing performance improvements solely due to a higher number of optimization epochs. The comparison of time complexities and corresponding performance is shown in Table \ref{complexity}.
Additionally, we adopt the same setup in our evaluation and sample the synthetic dataset after optimizing for 100 epochs in all of these different feature domains (i.e., $\mathbf{z}_{i}^{K-1}$). This approach prevents performance improvements obtained from implicitly selecting a dataset with a higher quality.

\begin{table}[!htbp]
    \centering
    \renewcommand{\arraystretch}{1.2}
    \resizebox{\linewidth}{!}{
    \begin{tabular}{lcccc}
        \toprule
        Commponent & ImNet-B  & ImNet-C  & ImWoof & ImNet-Fruits\\
        \midrule
        GLaD-TESLA            
        & 51.9\small{$\pm1.3$}     & 44.9\small{$\pm0.4$}     & 23.4\small{$\pm1.1$}  & 23.1\small{$\pm0.4$}\\
        + Average Initialization   & 53.5\small{$\pm0.7$}     & 46.1\small{$\pm0.9$}      & 24.8\small{$\pm1.1$}        & 22.7{\small{$\pm1.2$}}\textcolor{red}{$\downarrow$}\\
        + Hierarchical Layers    & 56.2\small{$\pm0.7$}     & 48.1\small{$\pm0.9$}      & 28.1\small{$\pm1.0$}         & 24.1\small{$\pm0.5$}\\
        \gc+ Distance Metric    & \gc \textbf{57.4}\small{$\pm0.3$}     & \gc \textbf{49.5}\small{$\pm0.6$}      & \gc \textbf{28.3}\small{$\pm0.2$}     & \gc \textbf{25.6}\small{$\pm0.7$}\\
        \midrule  
        GLaD-DSA  & 49.2\small{$\pm1.1$}     & 42.0\small{$\pm0.6$}     & 22.3\small{$\pm1.1$}        & 20.7\small{$\pm1.1$}\\
        + Average Initialization & 48.9{\small{$\pm0.8$}}\textcolor{red}{$\downarrow$}       & 40.6{\small{$\pm0.7$}}\textcolor{red}{$\downarrow$}     & 22.8\small{$\pm1.4$}    & 21.3\small{$\pm1.0$}\\
        + Hierarchical Layers    & 50.1\small{$\pm1.1$}     & 43.1\small{$\pm1.4$}      & 23.6\small{$\pm0.8$}          & 21.9\small{$\pm0.8$}\\
        \gc+ Distance Metric    & \gc \textbf{50.7}\small{$\pm0.9$}     & \gc \textbf{43.9}\small{$\pm0.7$}      & \gc \textbf{24.0}\small{$\pm0.8$}      & \gc \textbf{22.4}\small{$\pm1.1$}\\
        \midrule  
        GLaD-DM   & 42.9\small{$\pm1.9$}       & 39.4\small{$\pm1.7$}     & 21.2\small{$\pm1.5$}      & 21.8\small{$\pm1.8$}\\
        + Average Initialization   & 43.2\small{$\pm1.6$}     & 39.9\small{$\pm1.7$}     & 21.1{\small{$\pm1.9$}}\textcolor{red}{$\downarrow$}       & 22.3\small{$\pm1.3$}\\
        + Hierarchical Layers    & 44.2\small{$\pm2.1$}     & 41.0\small{$\pm1.2$}      & 23.1\small{$\pm0.9$}        & 24.1\small{$\pm1.4$}\\
        \gc+ Distance Metric    & \gc \textbf{44.7}\small{$\pm1.3$}     & \gc \textbf{41.1}\small{$\pm1.3$}      & \gc \textbf{23.9}\small{$\pm1.9$}      & \gc \textbf{24.4}\small{$\pm2.1$}\\
        \bottomrule
    \end{tabular}}
    \caption{Ablation study of each component with different distillation method across various ImageNet-Subset.}
    \label{component ablation}
    \vspace{-0.8em}
\end{table}
\subsection{Ablation Studies}
\label{exp:ablation}

\paragraph{Effectiveness of Each Component}

As Table \ref{component ablation} shows,
the two major components of our method, i.e., hierarchical feature domains and class-relevant feature distance 
both improve the performance across various ImageNet-Subsets with all distillation methods, especially on TESLA. 
Optimizing in an unfixed feature space can bring significant gains, and on this basis, using class-relevant feature distance for implicit evaluation can yield a slight additional improvement. Please note that using class-relevant feature distance is infeasible without unfixed optimization spaces.
Despite Initialization with averaged noise improves the performance of TESLA and DM to some degree, it cannot achieve stable improvement in the performance of the DSA method. We attribute this discrepancy to the inherent inclination towards noise and edge samples in DSA.
\begin{figure}[!htbp]
\centering
\resizebox{0.8\linewidth}{!}{
\includegraphics{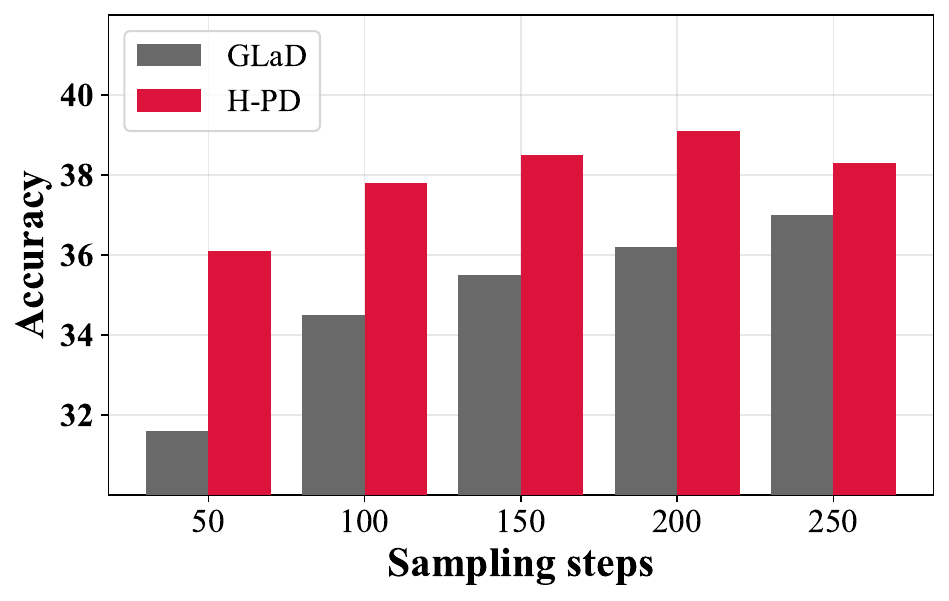}}
    \captionof{figure}{The comparison of performance(\%) at the same optimization epoch.}
    \label{align sampling}
\vspace{-0.1in}
\end{figure}
\begin{table}[!htbp]
    \centering
    \tabcolsep=0.6cm
    \resizebox{\linewidth}{!}{%
      \begin{tabular}{ccccc}
        \toprule
        Steps & TESLA  & DSA  & DM\\
        \midrule
        20 & 46.9\small{$\pm1.2$} & 39.8\small{$\pm1.1$} & \gc  \textbf{39.1}\small{$\pm1.2$} \\
        50 & 47.2\small{$\pm0.8$} & 41.6\small{$\pm0.8$} & 37.0\small{$\pm1.7$} \\
        100 & 50.3\small{$\pm0.6$} &\gc \textbf{43.2}\small{$\pm0.6$} & 36.5\small{$\pm1.4$} \\
        200 & \gc \textbf{50.5}\small{$\pm0.4$} & 43.0\small{$\pm0.6$} & 35.8\small{$\pm1.1$} \\
        \bottomrule
    \end{tabular}}
    \caption{Ablation results of optimization steps per optimization space averaged on ImageNet-[A, B, C, D, E].}
    \label{steps per layer}
    \vspace{-0.8em}
\end{table}
\paragraph{Optimization Steps}
\label{optimization steps}

We perform 100 optimization steps in each layer to align with the sampling method in previous works. We conduct more experiments across different optimization steps to explore the correct optimization steps. By observing the results shown in Table \ref{steps per layer}, we find that optimization steps beyond 100 do not yield significant performance improvements for TESLA and DSA methods. Meanwhile, optimization steps below 100 result in performance degradation. Considering the trade-off between effects and costs, we set the steps at 100 for both TESLA and DSA. For DM, however, optimal performance is attained at the least number of steps per layer, i.e., 20 steps. We compare the performance of GLaD and H-PD at the same epoch as shown in Figure \ref{align sampling}, our method outperforms and converges faster, demonstrating the superiority of the approach in utilizing the hierarchical features.
\begin{figure}[!htbp]
\centering
\resizebox{0.8\linewidth}{!}{
\includegraphics[width=\textwidth]{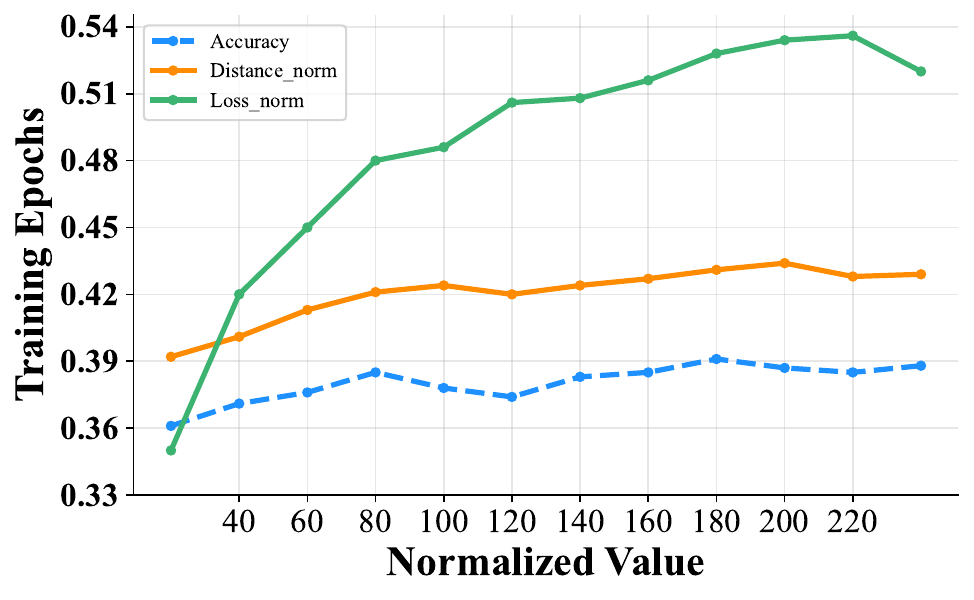}}
        \caption{The relationship between searching basis and performance. Note that higher loss-norm values indicate lower loss values and the same applies to feature distances.}
        \label{searching basis}
\vspace{-1.2em}
\end{figure}

\paragraph{Searching Basis}
To avoid the time-consuming task of directly evaluating the synthetic dataset, we opt for class-relevant feature distance for implicit searching. Specifically, we evaluate the synthetic dataset at specific epochs during the optimization process and subsequently record its ground-truth performance, loss value, and feature distance. Figure \ref{searching basis} demonstrates the recorded values during the optimization process. We normalize the loss values and feature distance to range $[0,1]$ for clear clarity and comparison. Our observation indicates that compared with the loss value, the feature distance consistently exhibits a stronger negative correlation with the  performance.

\section{Conclusion}
In this paper, we present a novel approach to dataset distillation by exploring hierarchical parameterization space and successfully enhance the GAN-based parameterization method. Our method transforms the optimization space from a specific GAN feature domain to a broader feature space, addressing challenges seen in previous GAN-based parameterization methods. An advantage is that our approach provides a new insight for parameterization methods. Additionally, we anticipate that further improvements can be achieved through detailed optimization steps and optimization space combinations. The proposed H-PD re-explores and showcases the potential of hierarchical features in parameterization distillation for enhancing the performance under extreme compression ratios, contributing to an advanced dataset distillation approach.\\
\textbf{Acknowledgement.} This work is supported in part by the National Natural Science Foundation of China under grant 62171248, 62301189, Peng Cheng Laboratory (PCL2023A08), and Shenzhen Science and Technology Program under Grant KJZD20240903103702004, JCYJ20220818101012025, RCBS20221008093124061, GXWD20220811172936001.
{
    \small
    \bibliographystyle{ieeenat_fullname}
    \bibliography{main}

\begin{thebibliography}{56}
\providecommand{\natexlab}[1]{#1}
\providecommand{\url}[1]{\texttt{#1}}
\expandafter\ifx\csname urlstyle\endcsname\relax
  \providecommand{\doi}[1]{doi: #1}\else
  \providecommand{\doi}{doi: \begingroup \urlstyle{rm}\Url}\fi

\bibitem[Bohdal et~al.(2020)Bohdal, Yang, and Hospedales]{bohdal2020flexible}
Ondrej Bohdal, Yongxin Yang, and Timothy Hospedales.
\newblock Flexible dataset distillation: Learn labels instead of images.
\newblock \emph{arXiv preprint arXiv:2006.08572}, 2020.

\bibitem[Brock et~al.(2016)Brock, Lim, Ritchie, and Weston]{brock2016neural}
Andrew Brock, Theodore Lim, James~M Ritchie, and Nick Weston.
\newblock Neural photo editing with introspective adversarial networks.
\newblock \emph{arXiv preprint arXiv:1609.07093}, 2016.

\bibitem[Brown et~al.(2020)Brown, Mann, Ryder, Subbiah, Kaplan, Dhariwal, Neelakantan, Shyam, Sastry, Askell, et~al.]{brown2020language}
Tom Brown, Benjamin Mann, Nick Ryder, Melanie Subbiah, Jared~D Kaplan, Prafulla Dhariwal, Arvind Neelakantan, Pranav Shyam, Girish Sastry, Amanda Askell, et~al.
\newblock Language models are few-shot learners.
\newblock \emph{Advances in neural information processing systems}, 33:\penalty0 1877--1901, 2020.

\bibitem[Cazenavette et~al.(2022)Cazenavette, Wang, Torralba, Efros, and Zhu]{cazenavette2022dataset}
George Cazenavette, Tongzhou Wang, Antonio Torralba, Alexei~A Efros, and Jun-Yan Zhu.
\newblock Dataset distillation by matching training trajectories.
\newblock In \emph{Proceedings of the IEEE/CVF Conference on Computer Vision and Pattern Recognition}, pages 4750--4759, 2022.

\bibitem[Cazenavette et~al.(2023)Cazenavette, Wang, Torralba, Efros, and Zhu]{cazenavette2023generalizing}
George Cazenavette, Tongzhou Wang, Antonio Torralba, Alexei~A Efros, and Jun-Yan Zhu.
\newblock Generalizing dataset distillation via deep generative prior.
\newblock In \emph{Proceedings of the IEEE/CVF Conference on Computer Vision and Pattern Recognition}, pages 3739--3748, 2023.

\bibitem[Chai et~al.(2021)Chai, Wulff, and Isola]{chai2021using}
Lucy Chai, Jonas Wulff, and Phillip Isola.
\newblock Using latent space regression to analyze and leverage compositionality in gans.
\newblock \emph{arXiv preprint arXiv:2103.10426}, 2021.

\bibitem[Chen et~al.(2025)Chen, Du, Huang, Wang, Zhang, and Wang]{cheninfluence}
Mingyang Chen, Jiawei Du, Bo Huang, Yi Wang, Xiaobo Zhang, and Wei Wang.
\newblock Influence-guided diffusion for dataset distillation.
\newblock In \emph{The Thirteenth International Conference on Learning Representations}, 2025.

\bibitem[Creswell et~al.(2018)Creswell, White, Dumoulin, Arulkumaran, Sengupta, and Bharath]{creswell2018generative}
Antonia Creswell, Tom White, Vincent Dumoulin, Kai Arulkumaran, Biswa Sengupta, and Anil~A Bharath.
\newblock Generative adversarial networks: An overview.
\newblock \emph{IEEE signal processing magazine}, 35\penalty0 (1):\penalty0 53--65, 2018.

\bibitem[Cui et~al.(2022)Cui, Wang, Si, and Hsieh]{cui2022dc}
Justin Cui, Ruochen Wang, Si Si, and Cho-Jui Hsieh.
\newblock Dc-bench: Dataset condensation benchmark.
\newblock \emph{Advances in Neural Information Processing Systems}, 35:\penalty0 810--822, 2022.

\bibitem[Cui et~al.(2023)Cui, Wang, Si, and Hsieh]{cui2023scaling}
Justin Cui, Ruochen Wang, Si Si, and Cho-Jui Hsieh.
\newblock Scaling up dataset distillation to imagenet-1k with constant memory.
\newblock In \emph{International Conference on Machine Learning}, pages 6565--6590. PMLR, 2023.

\bibitem[Deng et~al.(2009)Deng, Dong, Socher, Li, Li, and Fei-Fei]{deng2009imagenet}
Jia Deng, Wei Dong, Richard Socher, Li-Jia Li, Kai Li, and Li Fei-Fei.
\newblock Imagenet: A large-scale hierarchical image database.
\newblock In \emph{2009 IEEE conference on computer vision and pattern recognition}, pages 248--255. Ieee, 2009.

\bibitem[Devlin et~al.(2018)Devlin, Chang, Lee, and Toutanova]{devlin2018bert}
Jacob Devlin, Ming-Wei Chang, Kenton Lee, and Kristina Toutanova.
\newblock Bert: Pre-training of deep bidirectional transformers for language understanding.
\newblock \emph{arXiv preprint arXiv:1810.04805}, 2018.

\bibitem[Dosovitskiy et~al.(2020)Dosovitskiy, Beyer, Kolesnikov, Weissenborn, Zhai, Unterthiner, Dehghani, Minderer, Heigold, Gelly, et~al.]{dosovitskiy2020image}
Alexey Dosovitskiy, Lucas Beyer, Alexander Kolesnikov, Dirk Weissenborn, Xiaohua Zhai, Thomas Unterthiner, Mostafa Dehghani, Matthias Minderer, Georg Heigold, Sylvain Gelly, et~al.
\newblock An image is worth 16x16 words: Transformers for image recognition at scale.
\newblock \emph{arXiv preprint arXiv:2010.11929}, 2020.

\bibitem[Fang et~al.(2023)Fang, Chen, Wang, Wang, and Xia]{fang2023gifd}
Hao Fang, Bin Chen, Xuan Wang, Zhi Wang, and Shu-Tao Xia.
\newblock Gifd: A generative gradient inversion method with feature domain optimization.
\newblock In \emph{Proceedings of the IEEE/CVF International Conference on Computer Vision}, pages 4967--4976, 2023.

\bibitem[Fang et~al.(2024{\natexlab{a}})Fang, Kong, Yu, Chen, Li, Xia, and Xu]{fang2024one}
Hao Fang, Jiawei Kong, Wenbo Yu, Bin Chen, Jiawei Li, Shutao Xia, and Ke Xu.
\newblock One perturbation is enough: On generating universal adversarial perturbations against vision-language pre-training models.
\newblock \emph{arXiv preprint arXiv:2406.05491}, 2024{\natexlab{a}}.

\bibitem[Fang et~al.(2024{\natexlab{b}})Fang, Qiu, Yu, Yu, Kong, Chong, Chen, Wang, and Xia]{fang2024privacy}
Hao Fang, Yixiang Qiu, Hongyao Yu, Wenbo Yu, Jiawei Kong, Baoli Chong, Bin Chen, Xuan Wang, and Shu-Tao Xia.
\newblock Privacy leakage on dnns: A survey of model inversion attacks and defenses.
\newblock \emph{arXiv preprint arXiv:2402.04013}, 2024{\natexlab{b}}.

\bibitem[Gidaris and Komodakis(2018)]{gidaris2018dynamic}
Spyros Gidaris and Nikos Komodakis.
\newblock Dynamic few-shot visual learning without forgetting.
\newblock In \emph{Proceedings of the IEEE conference on computer vision and pattern recognition}, pages 4367--4375, 2018.

\bibitem[Gu et~al.(2024)Gu, Vahidian, Kungurtsev, Wang, Jiang, You, and Chen]{gu2024efficient}
Jianyang Gu, Saeed Vahidian, Vyacheslav Kungurtsev, Haonan Wang, Wei Jiang, Yang You, and Yiran Chen.
\newblock Efficient dataset distillation via minimax diffusion.
\newblock In \emph{Proceedings of the IEEE/CVF Conference on Computer Vision and Pattern Recognition}, pages 15793--15803, 2024.

\bibitem[He et~al.(2016)He, Zhang, Ren, and Sun]{he2016deep}
Kaiming He, Xiangyu Zhang, Shaoqing Ren, and Jian Sun.
\newblock Deep residual learning for image recognition.
\newblock In \emph{Proceedings of the IEEE conference on computer vision and pattern recognition}, pages 770--778, 2016.

\bibitem[Howard(2019)]{howard2019smaller}
Jeremy Howard.
\newblock A smaller subset of 10 easily classified classes from imagenet, and a little more french.
\newblock \emph{URL https://github. com/fastai/imagenette}, 2019.

\bibitem[Kim et~al.(2022)Kim, Kim, Oh, Yun, Song, Jeong, Ha, and Song]{kim2022dataset}
Jang-Hyun Kim, Jinuk Kim, Seong~Joon Oh, Sangdoo Yun, Hwanjun Song, Joonhyun Jeong, Jung-Woo Ha, and Hyun~Oh Song.
\newblock Dataset condensation via efficient synthetic-data parameterization.
\newblock In \emph{International Conference on Machine Learning}, pages 11102--11118. PMLR, 2022.

\bibitem[Krizhevsky(2009)]{krizhevsky2009learning}
A Krizhevsky.
\newblock Learning multiple layers of features from tiny images.
\newblock \emph{Master's thesis, University of Tront}, 2009.

\bibitem[Krizhevsky et~al.(2012)Krizhevsky, Sutskever, and Hinton]{krizhevsky2012imagenet}
Alex Krizhevsky, Ilya Sutskever, and Geoffrey~E Hinton.
\newblock Imagenet classification with deep convolutional neural networks.
\newblock \emph{Advances in neural information processing systems}, 25, 2012.

\bibitem[Le and Yang(2015)]{le2015tiny}
Ya Le and Xuan Yang.
\newblock Tiny imagenet visual recognition challenge.
\newblock \emph{CS 231N}, 7\penalty0 (7):\penalty0 3, 2015.

\bibitem[Lee et~al.(2022)Lee, Chun, Jung, Yun, and Yoon]{lee2022dataset}
Saehyung Lee, Sanghyuk Chun, Sangwon Jung, Sangdoo Yun, and Sungroh Yoon.
\newblock Dataset condensation with contrastive signals.
\newblock In \emph{International Conference on Machine Learning}, pages 12352--12364. PMLR, 2022.

\bibitem[Lei and Tao(2023)]{lei2023comprehensive}
Shiye Lei and Dacheng Tao.
\newblock A comprehensive survey to dataset distillation.
\newblock \emph{arXiv preprint arXiv:2301.05603}, 2023.

\bibitem[Liu et~al.(2022)Liu, Wang, Yang, Ye, and Wang]{liu2022dataset}
Songhua Liu, Kai Wang, Xingyi Yang, Jingwen Ye, and Xinchao Wang.
\newblock Dataset distillation via factorization.
\newblock \emph{Advances in Neural Information Processing Systems}, 35:\penalty0 1100--1113, 2022.

\bibitem[Liu et~al.(2021)Liu, Yamada, Tsai, Le, Salakhutdinov, and Yang]{liu2021lsmi}
Yanbin Liu, Makoto Yamada, Yao-Hung~Hubert Tsai, Tam Le, Ruslan Salakhutdinov, and Yi Yang.
\newblock Lsmi-sinkhorn: Semi-supervised mutual information estimation with optimal transport.
\newblock In \emph{Machine Learning and Knowledge Discovery in Databases. Research Track: European Conference, ECML PKDD 2021, Bilbao, Spain, September 13--17, 2021, Proceedings, Part I 21}, pages 655--670. Springer, 2021.

\bibitem[Liu et~al.(2023)Liu, Gu, Wang, Zhu, Jiang, and You]{liu2023dream}
Yanqing Liu, Jianyang Gu, Kai Wang, Zheng Zhu, Wei Jiang, and Yang You.
\newblock Dream: Efficient dataset distillation by representative matching.
\newblock \emph{arXiv preprint arXiv:2302.14416}, 2023.

\bibitem[Muhammad and Yeasin(2020)]{muhammad2020eigen}
Mohammed~Bany Muhammad and Mohammed Yeasin.
\newblock Eigen-cam: Class activation map using principal components.
\newblock In \emph{2020 international joint conference on neural networks (IJCNN)}, pages 1--7. IEEE, 2020.

\bibitem[Nguyen et~al.(2020)Nguyen, Chen, and Lee]{nguyen2020dataset}
Timothy Nguyen, Zhourong Chen, and Jaehoon Lee.
\newblock Dataset meta-learning from kernel ridge-regression.
\newblock \emph{arXiv preprint arXiv:2011.00050}, 2020.

\bibitem[Nguyen et~al.(2021)Nguyen, Novak, Xiao, and Lee]{nguyen2021dataset}
Timothy Nguyen, Roman Novak, Lechao Xiao, and Jaehoon Lee.
\newblock Dataset distillation with infinitely wide convolutional networks.
\newblock \emph{Advances in Neural Information Processing Systems}, 34:\penalty0 5186--5198, 2021.

\bibitem[Qiu et~al.(2024)Qiu, Fang, Yu, Chen, Qiu, and Xia]{qiu2024closer}
Yixiang Qiu, Hao Fang, Hongyao Yu, Bin Chen, MeiKang Qiu, and Shu-Tao Xia.
\newblock A closer look at gan priors: Exploiting intermediate features for enhanced model inversion attacks.
\newblock In \emph{European Conference on Computer Vision}, pages 109--126. Springer, 2024.

\bibitem[Rombach et~al.(2022)Rombach, Blattmann, Lorenz, Esser, and Ommer]{rombach2022high}
Robin Rombach, Andreas Blattmann, Dominik Lorenz, Patrick Esser, and Bj{\"o}rn Ommer.
\newblock High-resolution image synthesis with latent diffusion models.
\newblock In \emph{Proceedings of the IEEE/CVF conference on computer vision and pattern recognition}, pages 10684--10695, 2022.

\bibitem[Sajedi et~al.(2023)Sajedi, Khaki, Amjadian, Liu, Lawryshyn, and Plataniotis]{sajedi2023datadam}
Ahmad Sajedi, Samir Khaki, Ehsan Amjadian, Lucy~Z Liu, Yuri~A Lawryshyn, and Konstantinos~N Plataniotis.
\newblock Datadam: Efficient dataset distillation with attention matching.
\newblock In \emph{Proceedings of the IEEE/CVF International Conference on Computer Vision}, pages 17097--17107, 2023.

\bibitem[Sauer et~al.(2022)Sauer, Schwarz, and Geiger]{sauer2022stylegan}
Axel Sauer, Katja Schwarz, and Andreas Geiger.
\newblock Stylegan-xl: Scaling stylegan to large diverse datasets.
\newblock In \emph{ACM SIGGRAPH 2022 conference proceedings}, pages 1--10, 2022.

\bibitem[Shen and Xing(2022)]{shen2022fast}
Zhiqiang Shen and Eric Xing.
\newblock A fast knowledge distillation framework for visual recognition.
\newblock In \emph{European conference on computer vision}, pages 673--690. Springer, 2022.

\bibitem[Shin et~al.(2023)Shin, Shin, and Moon]{shin2023frequency}
DongHyeok Shin, Seungjae Shin, and Il-chul Moon.
\newblock Frequency domain-based dataset distillation.
\newblock In \emph{Thirty-seventh Conference on Neural Information Processing Systems}, 2023.

\bibitem[Simonyan and Zisserman(2014)]{simonyan2014very}
Karen Simonyan and Andrew Zisserman.
\newblock Very deep convolutional networks for large-scale image recognition.
\newblock \emph{arXiv preprint arXiv:1409.1556}, 2014.

\bibitem[Su et~al.(2024)Su, Hou, Gao, Tian, and Tang]{su2024d}
Duo Su, Junjie Hou, Weizhi Gao, Yingjie Tian, and Bowen Tang.
\newblock D\^{} 4: Dataset distillation via disentangled diffusion model.
\newblock In \emph{Proceedings of the IEEE/CVF Conference on Computer Vision and Pattern Recognition}, pages 5809--5818, 2024.

\bibitem[Tewari et~al.(2020)Tewari, Elgharib, Bernard, Seidel, P{\'e}rez, Zollh{\"o}fer, and Theobalt]{tewari2020pie}
Ayush Tewari, Mohamed Elgharib, Florian Bernard, Hans-Peter Seidel, Patrick P{\'e}rez, Michael Zollh{\"o}fer, and Christian Theobalt.
\newblock Pie: Portrait image embedding for semantic control.
\newblock \emph{ACM Transactions on Graphics (TOG)}, 39\penalty0 (6):\penalty0 1--14, 2020.

\bibitem[Ulyanov et~al.(2016)Ulyanov, Vedaldi, and Lempitsky]{ulyanov2016instance}
Dmitry Ulyanov, Andrea Vedaldi, and Victor Lempitsky.
\newblock Instance normalization: The missing ingredient for fast stylization.
\newblock \emph{arXiv preprint arXiv:1607.08022}, 2016.

\bibitem[Wang et~al.(2022)Wang, Zhao, Peng, Zhu, Yang, Wang, Huang, Bilen, Wang, and You]{wang2022cafe}
Kai Wang, Bo Zhao, Xiangyu Peng, Zheng Zhu, Shuo Yang, Shuo Wang, Guan Huang, Hakan Bilen, Xinchao Wang, and Yang You.
\newblock Cafe: Learning to condense dataset by aligning features.
\newblock In \emph{Proceedings of the IEEE/CVF Conference on Computer Vision and Pattern Recognition}, pages 12196--12205, 2022.

\bibitem[Wang et~al.(2018)Wang, Zhu, Torralba, and Efros]{wang2018dataset}
Tongzhou Wang, Jun-Yan Zhu, Antonio Torralba, and Alexei~A Efros.
\newblock Dataset distillation.
\newblock \emph{arXiv preprint arXiv:1811.10959}, 2018.

\bibitem[Xia et~al.(2022)Xia, Zhang, Yang, Xue, Zhou, and Yang]{xia2022gan}
Weihao Xia, Yulun Zhang, Yujiu Yang, Jing-Hao Xue, Bolei Zhou, and Ming-Hsuan Yang.
\newblock Gan inversion: A survey.
\newblock \emph{IEEE Transactions on Pattern Analysis and Machine Intelligence}, 45\penalty0 (3):\penalty0 3121--3138, 2022.

\bibitem[Yin et~al.(2024)Yin, Xing, and Shen]{yin2024squeeze}
Zeyuan Yin, Eric Xing, and Zhiqiang Shen.
\newblock Squeeze, recover and relabel: Dataset condensation at imagenet scale from a new perspective.
\newblock \emph{Advances in Neural Information Processing Systems}, 36, 2024.

\bibitem[Zhang et~al.(2017)Zhang, Cisse, Dauphin, and Lopez-Paz]{zhang2017mixup}
Hongyi Zhang, Moustapha Cisse, Yann~N Dauphin, and David Lopez-Paz.
\newblock mixup: Beyond empirical risk minimization.
\newblock \emph{arXiv preprint arXiv:1710.09412}, 2017.

\bibitem[Zhao and Bilen(2021)]{zhao2021dataset}
Bo Zhao and Hakan Bilen.
\newblock Dataset condensation with differentiable siamese augmentation.
\newblock In \emph{International Conference on Machine Learning}, pages 12674--12685. PMLR, 2021.

\bibitem[Zhao and Bilen(2022)]{zhao2022synthesizing}
Bo Zhao and Hakan Bilen.
\newblock Synthesizing informative training samples with gan.
\newblock \emph{arXiv preprint arXiv:2204.07513}, 2022.

\bibitem[Zhao and Bilen(2023)]{zhao2023dataset}
Bo Zhao and Hakan Bilen.
\newblock Dataset condensation with distribution matching.
\newblock In \emph{Proceedings of the IEEE/CVF Winter Conference on Applications of Computer Vision}, pages 6514--6523, 2023.

\bibitem[Zhao et~al.(2020)Zhao, Mopuri, and Bilen]{zhao2020dataset}
Bo Zhao, Konda~Reddy Mopuri, and Hakan Bilen.
\newblock Dataset condensation with gradient matching.
\newblock \emph{arXiv preprint arXiv:2006.05929}, 2020.

\bibitem[Zhong et~al.(2024{\natexlab{a}})Zhong, Chen, Fang, Gu, Xia, and Yang]{zhong2024going}
Xinhao Zhong, Bin Chen, Hao Fang, Xulin Gu, Shu-Tao Xia, and En-Hui Yang.
\newblock Going beyond feature similarity: Effective dataset distillation based on class-aware conditional mutual information.
\newblock \emph{arXiv preprint arXiv:2412.09945}, 2024{\natexlab{a}}.

\bibitem[Zhong et~al.(2024{\natexlab{b}})Zhong, Sun, Gu, Xu, Wang, Wu, and Chen]{zhong2024efficient}
Xinhao Zhong, Shuoyang Sun, Xulin Gu, Zhaoyang Xu, Yaowei Wang, Jianlong Wu, and Bin Chen.
\newblock Efficient dataset distillation via diffusion-driven patch selection for improved generalization.
\newblock \emph{arXiv preprint arXiv:2412.09959}, 2024{\natexlab{b}}.

\bibitem[Zhou et~al.(2022)Zhou, Nezhadarya, and Ba]{zhou2022dataset}
Yongchao Zhou, Ehsan Nezhadarya, and Jimmy Ba.
\newblock Dataset distillation using neural feature regression.
\newblock \emph{Advances in Neural Information Processing Systems}, 35:\penalty0 9813--9827, 2022.

\bibitem[Zhu et~al.(2024{\natexlab{a}})Zhu, Li, Ma, He, and Li]{zhu2024multibooth}
Chenyang Zhu, Kai Li, Yue Ma, Chunming He, and Xiu Li.
\newblock Multibooth: Towards generating all your concepts in an image from text.
\newblock \emph{arXiv preprint arXiv:2404.14239}, 2024{\natexlab{a}}.

\bibitem[Zhu et~al.(2024{\natexlab{b}})Zhu, Li, Ma, Tang, Fang, Chen, Chen, and Li]{zhu2024instantswap}
Chenyang Zhu, Kai Li, Yue Ma, Longxiang Tang, Chengyu Fang, Chubin Chen, Qifeng Chen, and Xiu Li.
\newblock Instantswap: Fast customized concept swapping across sharp shape differences.
\newblock \emph{arXiv preprint arXiv:2412.01197}, 2024{\natexlab{b}}.

\end{thebibliography}
}

\clearpage
\maketitlesupplementary
\setcounter{section}{0}
\renewcommand{\thesection}{\Alph{section}}



\section{Literature Reviews on Dataset Distillation}
\label{appendix:literature reviews}

\subsection{Dataset Distillation in Pixel Space}
In this section, we review the methodology of optimizing synthetic dataset $S$ with the surrogate objective in pixel space, which provides the basic optimization objective for all parameterization dataset distillation methods.

\subsubsection{DC \cite{zhao2020dataset}.} Dataset Distillation (DD) \cite{wang2018dataset} aims at optimizing the synthetic dataset $\mathcal{S}$ with a bi-level optimization. The main idea of bi-level optimization is that a network with parameter $\theta_{\mathcal{S}}$, which is trained on $\mathcal{S}$, should minimize the risk of the real dataset $\mathcal{T}$. However, due to the need to pass through an unrolled computation graph, DD brings about a significant amount of time overhead. Based on this, DC introduces a surrogate objective, which aims at matching the gradients of a network during the optimization. For a network with parameters $\theta_{S}$ trained on the synthetic data for some number of iterations, the matching loss is
\begin{eqnarray}
\mathcal{L}_{\mathrm{DC}}=1-\frac{\nabla_{\theta} \ell^{\mathcal{S}}(\theta) \cdot \nabla_{\theta} \ell^{\mathcal{T}}(\theta)}{\left\|\nabla_{\theta} \ell^{\mathcal{S}}(\theta)\right\|\left\|\nabla_{\theta} \ell^{\mathcal{T}}(\theta)\right\|} ,
\end{eqnarray}

where $\ell^{\mathcal{T}}(\cdot)$ represents the loss function (e.g., CE loss) calculated on real dataset $\mathcal{T}$, and $\ell^{\mathcal{S}}(\cdot)$ is the same loss function calculated on synthetic dataset $\mathcal{T}$.

\subsubsection{DM \cite{zhao2023dataset}.} Despite DC significantly reducing time consumption through surrogate, bi-level optimization still introduces a substantial amount of time overhead, especially when dealing with high-resolution and large-scale datasets. DM achieves this by using only the features extracted from networks $\psi$ with random initialization as the matching target, the matching loss is
\begin{eqnarray}
\mathcal{L}_{\mathrm{DM}}=\sum_{c}\left\|\frac{1}{\left|\mathcal{T}_{c}\right|} \sum_{\mathbf{x} \in \mathcal{T}_{c}} \psi(\mathbf{x})-\frac{1}{\left|\mathcal{S}_{c}\right|} \sum_{\mathbf{s} \in \mathcal{S}_{c}} \psi(\mathbf{s})\right\|^{2} ,
\end{eqnarray}

where $\mathcal{T}_{c}$ and $\mathcal{S}_{c}$ represents the real and synthetic images from class $c$ respectively.

\subsubsection{MTT \cite{cazenavette2022dataset}.} Distinct from the short-range optimization introduced from DC, MTT utilizes many expert trajectories $\{\theta^{*}_{t}\}^{T}_{0}$ which are obtained by training networks from scratch on the full real dataset and choose the parameter distance the matching objective.  
During the distillation process, a student network is initialized with parameters $\theta^{*}_{t}$  by sample expert trajectory at timestamp $t$ and then trained on the synthetic data for some number of iterations $N$, the matching loss is 
\begin{eqnarray}
\mathcal{L}_{\mathrm{MTT}}=\frac{\left\|\hat{\theta}_{t+N}-\theta_{t+M}^{*}\right\|^{2}}{\left\|\theta_{t}^{*}-\theta_{t+M}^{*}\right\|^{2}} ,
\end{eqnarray}

where $\theta_{t+M}^{*}$ represents the expert trajectory at timestamp $t+M$.

\subsection{Dataset Distillation in Feature Domain}
In this section, we review the methodology of parameterization dataset distillation built upon the aforementioned dataset distillation methods, achieving better performance by employing a differentiable operation $\mathcal{F}(\cdot)$ to shift the optimization space from pixel space to various feature domain, which can be formulated as
\begin{eqnarray}
\mathcal{S} = \{\mathcal{F}(\mathbf{z})\} .
\end{eqnarray}

where $\mathbf{z}$ represents latent code in the feature domain corresponding to $\mathcal{F}(\cdot)$.

\subsubsection{HaBa \cite{liu2022dataset}.} 
HaBa breaks the synthetic dataset into bases and a small neural network called hallucinator which is utilized to produce additional synthetic images. By leveraging this technique, the resulting model could be regarded as a differentiable operation and produce more diverse samples. However, HaBa simultaneously optimizes the bases and the hallucinator, neglecting the relationship between the two feature domains. This leads to unstable optimization during the training process.

\subsubsection{IDC \cite{kim2022dataset}.} 
IDC proposes a principle that small-sized synthetic images often carry more effective information under the same spatial budget and utilize an upsampling module as the differentiable operation. Despite employing a differentiable operation, the optimization of IDC is still the pixel space, which resulted in the loss of effective information gain obtained from other feature domains.

\subsubsection{FreD \cite{shin2023frequency}.} 
FreD suggests that optimizing for the main subject in the synthetic image is more instructive than optimizing for all the details. Therefore, FreD employs discrete cosine transform (DCT) as the differentiable operation and uses a learnable mask matrix to remove high-frequency information, ensuring that the optimization process only occurs in the low-frequency domain. This allows the synthetic dataset to achieve higher performance and generalization. However, FreD overlooks the effective guiding information within the high-frequency domain and fails to connect the two feature domains produced by DCT, leading to potential incomplete optimization.

\subsubsection{GLaD \cite{cazenavette2023generalizing}.} Different from existing methods~\cite{gu2024efficient,su2024d,zhong2024efficient,cheninfluence} utilizing diffusion models~\cite{zhu2024instantswap,zhu2024multibooth}, GLaD employs a pre-trained generative model (i.e., GAN) and distills the synthetic dataset in the corresponding latent space. By leveraging the capability of a generative model to map latent noise to image patterns, GLaD achieves better generalization to unseen architecture and scale to high-dimensional datasets. However, for StyleGAN, the earlier layers tend to provide the information about the main subject in an image while the later layers often contribute to the details. However, GLaD attempts to balance the low-frequency information with the high-frequency information by selecting an intermediate layer as a fixed optimization space, discarding the guiding information from the earlier layers can lead to incomplete optimization. Another limitation of GLaD is the need for a large number of preliminary experiments. GLaD selects a specific intermediate layer suitable for all datasets for different distillation methods, However, under the same distillation method, the optimal intermediate layer corresponding to different datasets is not the same, especially when the manifold of the datasets varies greatly, which suggests that GLaD cannot spontaneously adapt to different datasets, distillation methods, and GANs.


\section{Additional Experimental Results} 
\label{additional results}

\begin{table*}[!htbp]
    \centering
    \renewcommand{\arraystretch}{1.2}
    \resizebox{\textwidth}{!}{
    \begin{tabular}{cccccccccccc}
        \toprule
        Alg. & Opimization Space & ImNet-A & ImNet-B & ImNet-C & ImNet-D & ImNet-E & ImNette & ImWoof & ImNet-Birds & ImNet-Fruits & ImNet-Cats\\
        \midrule
            & Fixed (Pixel)  & 51.7\small{$\pm0.2$}     & 53.3\small{$\pm1.0$}     & 48.0\small{$\pm0.7$}     &43.0\small{$\pm0.6$}      & 39.5\small{$\pm0.9$}     & 41.8\small{$\pm0.6$}     & 22.6\small{$\pm0.6$}    & 37.3\small{$\pm0.8$}       & 22.4\small{$\pm1.1$}       & 22.6\small{$\pm0.4$}\\
        TESLA & Fixed (GAN)  & 50.7\small{$\pm0.4$}     & 51.9\small{$\pm1.3$}     & 44.9\small{$\pm0.4$}     & 39.9\small{$\pm1.7$}     & 37.6\small{$\pm0.7$}     & 38.7\small{$\pm1.6$}     & 23.4\small{$\pm1.1$}    & 35.8\small{$\pm1.4$}       & 23.1\small{$\pm0.4$}       & 26.0\small{$\pm1.1$}\\
            & \gc Unfixed   &\gc \textbf{53.1}\small{$\pm0.8$}     &\gc \textbf{55.4}\small{$\pm0.7$}      &\gc \textbf{47.5}\small{$\pm0.9$}     &\gc \textbf{44.1}\small{$\pm0.6$}       &\gc \textbf{40.8}\small{$\pm0.7$}     &\gc \textbf{42.8}\small{$\pm1.0$}     &\gc \textbf{27.0}\small{$\pm0.6$}      &\gc \textbf{37.6}\small{$\pm0.9$}          &\gc \textbf{24.7}\small{$\pm0.7$}          &\gc \textbf{28.3}\small{$\pm0.8$}\\
        \midrule
            & Fixed (Pixel)  & 43.2\small{$\pm0.6$}   & 47.2\small{$\pm0.7$}     & 41.3\small{$\pm0.7$}     & 34.3\small{$\pm1.5$}       & 34.9\small{$\pm1.5$}     & 34.2\small{$\pm1.7$}     & 22.5\small{$\pm1.0$}     & 32.0\small{$\pm1.5$}    & 21.0\small{$\pm0.9$}      & 22.0\small{$\pm0.6$}  \\
        DSA  & Fixed (GAN)  & 44.1\small{$\pm2.4$}     & 49.2\small{$\pm1.1$}     & 42.0\small{$\pm0.6$}       & 35.6\small{$\pm0.9$}     & 35.8\small{$\pm0.9$}     & 35.4\small{$\pm1.2$}     & 22.3\small{$\pm1.1$}    & 33.8\small{$\pm0.9$}         & 20.7\small{$\pm1.1$}          & 22.6\small{$\pm0.8$}\\
            &\gc Unfixed   &\gc \textbf{46.1}\small{$\pm0.7$}      &\gc \textbf{50.0}\small{$\pm0.9$}     & \gc \textbf{43.8}\small{$\pm1.4$}     & \gc \textbf{37.1}\small{$\pm0.9$}     & \gc \textbf{36.6}\small{$\pm0.6$}     & \gc
            \textbf{36.2}\small{$\pm0.5$}     &\gc \textbf{22.7}\small{$\pm0.3$}    &\gc \textbf{34.9}\small{$\pm1.5$}           &\gc \textbf{21.2}\small{$\pm0.8$}          &\gc \textbf{23.1}\small{$\pm0.3$}\\
        \midrule
            & Fixed (Pixel)  & 39.4\small{$\pm1.8$}       & 40.9\small{$\pm1.7$}     & 39.0\small{$\pm1.3$}     & 30.8\small{$\pm0.9$}     & 27.0\small{$\pm0.8$}     & 30.4\small{$\pm2.7$}     & 20.7\small{$\pm1.0$}    & 26.6\small{$\pm2.6$}        & 20.4\small{$\pm1.9$}        & 20.1\small{$\pm1.2$}\\
        DM  & Fixed (GAN)  & 41.0\small{$\pm1.5$}       & 42.9\small{$\pm1.9$}     & 39.4\small{$\pm1.7$}     & 33.2\small{$\pm1.4$}     & 30.3\small{$\pm1.3$}     & 32.2\small{$\pm1.7$}     & 21.2\small{$\pm1.5$}    & 27.6\small{$\pm1.9$}         & 21.8\small{$\pm1.8$}          & 22.3\small{$\pm1.6$}\\
            &\gc Unfixed   &\gc \textbf{42.3}\small{$\pm1.5$}     &\gc \textbf{44.1}\small{$\pm1.5$}     &\gc \textbf{41.3}\small{$\pm1.7$}     & \gc\textbf{33.7}\small{$\pm1.1$}     &\gc \textbf{31.5}\small{$\pm1.1$}     &\gc \textbf{34.0}\small{$\pm1.2$}     &\gc \textbf{23.1}\small{$\pm1.3$}    &\gc \textbf{28.9}\small{$\pm1.4$}         &\gc \textbf{24.3}\small{$\pm1.3$}          &\gc \textbf{22.8}\small{$\pm0.8$}\\
        \bottomrule
    \end{tabular}}
    \caption{Abltion study on optimization space comparison. "Fixed (Pixel)" refers to optimize in pixel space and "Fixed (GAN)" refers to GLaD, while Unfixed refers to optimize in multiple feature domains.}
    \label{performance:fixed-unfixed}
\end{table*}

\subsection{More Comparisons with GLaD}
To expand the optimization space, the method we proposed utilizes hierarchical feature domains composed of intermediate layers from  GAN. To investigate whether optimization across multiple feature domains is superior to optimization within a single fixed feature domain, we evaluate the performance by simply expanding the optimization space based on the baseline. As shown in Table \ref{performance:fixed-unfixed}, compared to GLaD, which only selects a single yet optimal intermediate layer of the GAN as the optimization space, H-PD has successfully achieved considerable improvement, validating our viewpoint that the optimization result from the previous feature domain can serve as better starting point for subsequent feature domain. Please note the result is obtained by not selecting $\mathcal{S}^{*}$.
\begin{figure}[!htbp]
\centering
\resizebox{\linewidth}{!}{
\includegraphics{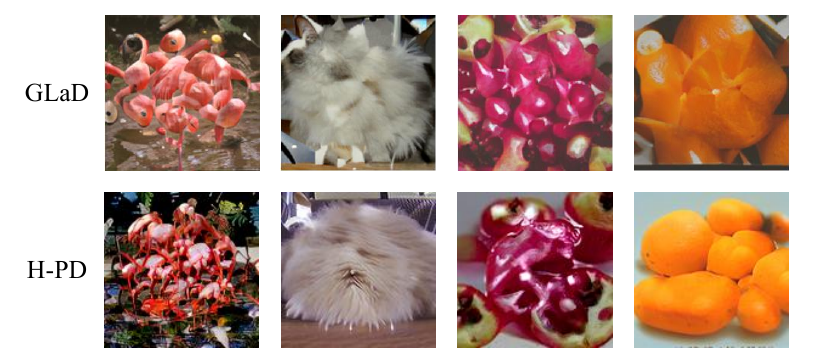}}
    \captionof{figure}{The comparison of visualization.}
    \label{visualization comparison}
\end{figure}

\begin{table}[!htbp]
    \centering
    \resizebox{\linewidth}{!}{%
    \begin{tabular}{cccccc}
        \toprule
        Method & ImNet-A & ImNet-B & ImNet-C & ImNet-D & ImNet-D \\
        \midrule
        Pixel & 38.3\small{$\pm4.7$} & 32.8\small{$\pm4.1$} & 27.6\small{$\pm3.3$} & 25.5\small{$\pm1.2$} & 23.5\small{$\pm2.4$} \\
        \midrule
        GLaD & 37.4\small{$\pm5.5$} & 41.5\small{$\pm1.2$} & 35.7\small{$\pm4.0$} & 27.9\small{$\pm1.0$} & 29.3\small{$\pm1.2$} \\
        \midrule
        \gc H-PD &\gc  \textbf{40.7}\small{$\pm2.1$} &\gc  \textbf{42.9}\small{$\pm1.8$} &\gc  \textbf{37.2}\small{$\pm2.2$} &\gc  \textbf{30.1}\small{$\pm1.7$} &\gc \textbf{29.7}\small{$\pm1.8$} \\
        \bottomrule
    \end{tabular}}
    \caption{Higher-resolution (256×256) synthetic dataset (using DSA) cross-architecture performance (\%).}
    \label{high-resolution}
\end{table}

To present a more comprehensive comparison, we evaluate the cross-architecture performance of a high-resolution synthetic dataset under the same setting (i.e., DSA on ImageNet-[A, B, C, D, E] under IPC=1).  As shown in Table \ref{high-resolution}, our proposed H-PD still achieves considerable improvements, demonstrating the stability of our proposed method. Figure \ref{visualization comparison} illustrates the comparison of synthetic dataset visualization generated by H-PD and GLaD using the same initial image. The images produced by H-PD achieve a good balance between content and style. On one hand, H-PD tends to preserve more main subject information by optimizing in the earlier layers of the GAN. On the other hand, since H-PD also undergoes optimization in the later layers, the synthetic images tend to be sharper and rarely produce the kaleidoscope-like patterns that are common in the GLaD method.

\begin{figure*}[!htbp]
\centering
	\subfloat[Visualization of synthetic images.]{
    \label{vis change}
    \includegraphics[width = 0.48\textwidth]{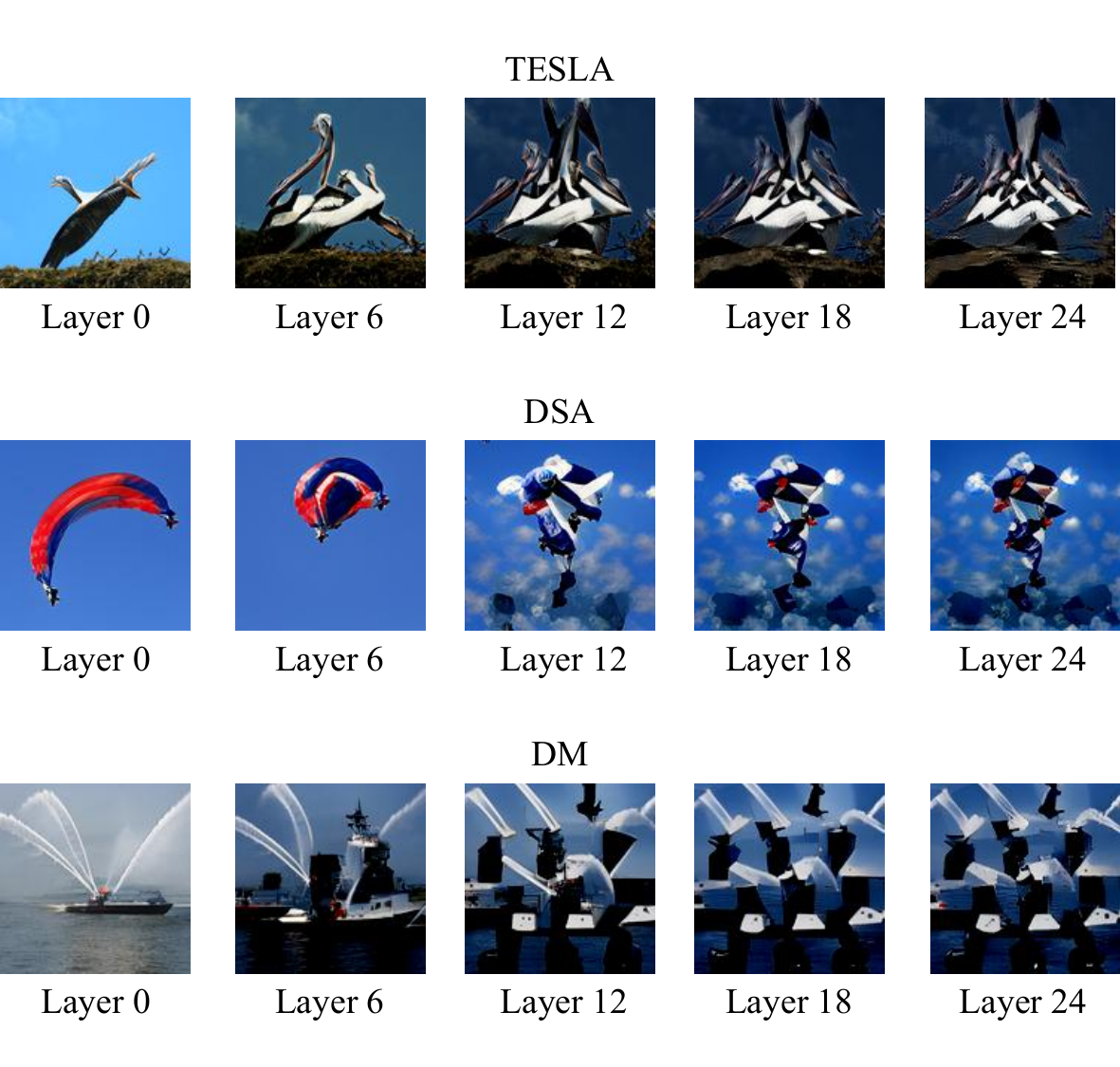}}
	\hfill
	\subfloat[Visualization of corresponding CAM.]{
    \label{cam change}
    \includegraphics[width = 0.48\textwidth]{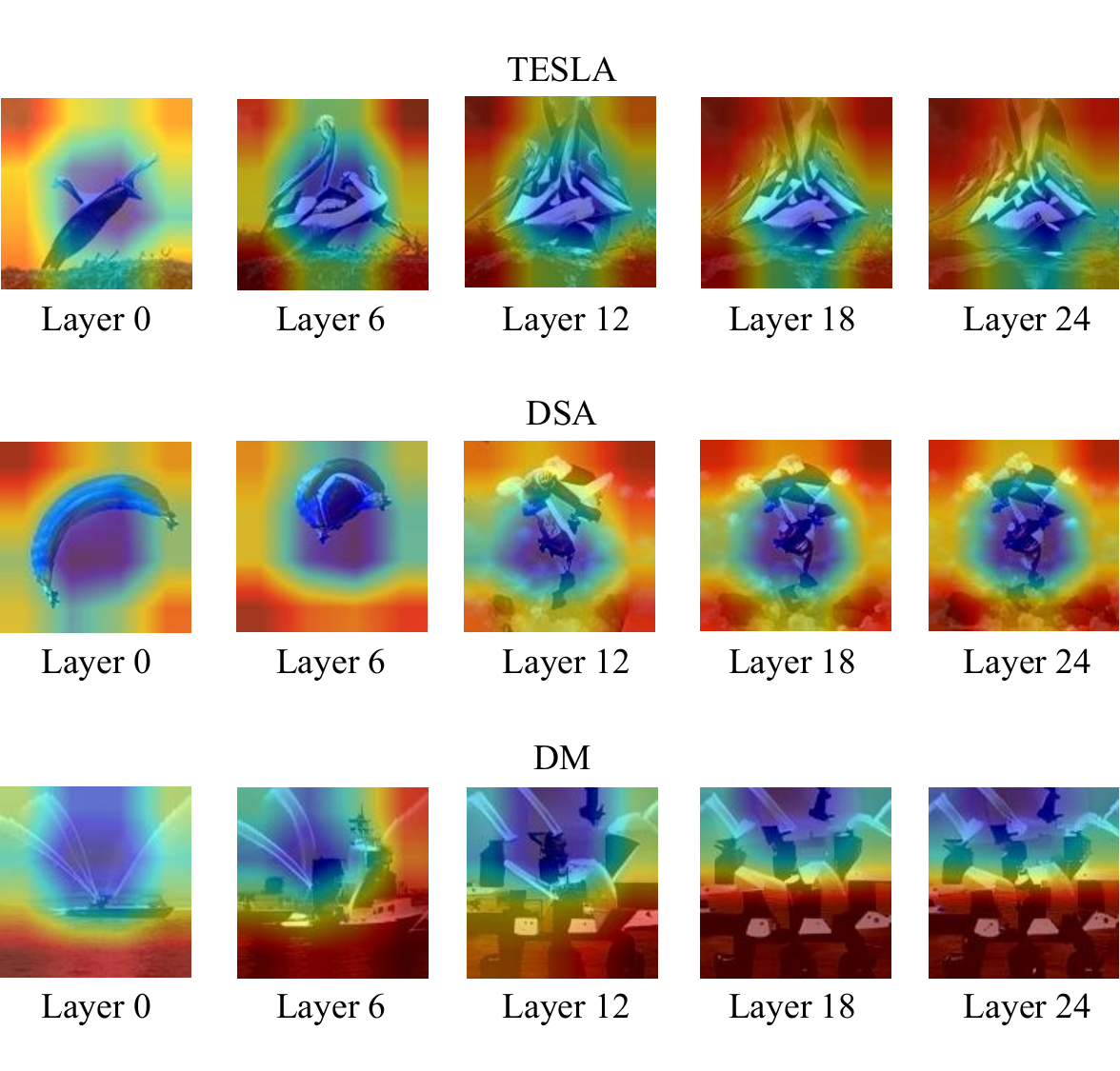}}
    \caption{The visualization change of synthetic images and corresponding CAM during the optimization process using different distillation methods. "Layer" refers to the index of intermediate layers provided by StyleGAN-XL.}
    \label{change}
\end{figure*}

\subsection{Visualizing Morphological Transition of Synthetic Images }
\label{qualitative interpretation}
As shown in Figure \ref{vis change}, we demonstrate the visualization changes of the synthetic image throughout the optimization process. Layer $0$ represents the initial image produced by StyleGAN-XL using averaged noise, and Layer $i$ indicates the image when the optimization space reaches layer $i$. In the early stage of optimization, since the optimization space is located in the earlier layer of the GAN, the optimization object primarily focus on the main subject of the synthetic image. Meanwhile, GAN still maintains a high degree of integrity which leads to a strong constraint on the slight changes in the latent produced during the optimization process, which can be transformed into patterns resembling real images instead of noises. Thus the tendency in the early stage of optimization is to generate images that better conform to the constraint of distillation loss yet appear more realistic, leading to produce synthetic images that can be regarded as a better starting point for the subsequent optimization process.

In the later stage of optimization, the main subject of the synthetic image no longer undergoes significant changes, and the optimization objective shifts along with the movement of the optimization space, focusing more on the details of the synthetic images. As shown in Figure \ref{vis change}, due to the weakened generative constraint of the incomplete GAN, the final synthetic image becomes similar to the indistinguishable and distorted image produced by existing distillation methods. Building upon the better synthetic image obtained through the optimization process in the earlier layers, different distillation methods gradually incorporate more guidance-oriented customized patterns into the synthetic image, achieving further performance improvement, which has also been proved by recent work~\cite{zhong2024going}.

\begin{table*}[!htbp]
    \centering
    \renewcommand{\arraystretch}{1.2}
    \resizebox{\textwidth}{!}{
    \begin{tabular}{cccccccccccc}
        \toprule
        Layers & Optimization & ImNet-A & ImNet-B & ImNet-C & ImNet-D & ImNet-E & ImNette & ImWoof & ImNet-Birds & ImNet-Fruits & ImNet-Cats\\
        \midrule
            & 50  & 53.6\small{$\pm0.2$}     & 55.2\small{$\pm1.5$}     
            & 47.3\small{$\pm0.5$}     &44.1\small{$\pm0.7$}      
            & 40.5\small{$\pm1.1$}     
            & 43.8\small{$\pm0.4$}     
            & 26.6\small{$\pm0.7$}    & 37.1\small{$\pm0.6$}       & 22.9\small{$\pm0.5$}       & 27.8\small{$\pm1.0$}\\
        1 & 100   & 55.3\small{$\pm0.8$}    
        & 57.1\small{$\pm0.7$}     
        &\rc 49.1\small{$\pm0.9$}     
        &\gc \textbf{46.6}\small{$\pm0.4$}    
        & 42.2\small{$\pm1.5$}     
        & 44.9\small{$\pm1.2$}     
        & 28.6\small{$\pm0.6$}    
        & 39.4\small{$\pm0.8$}       
        &\rc 25.9\small{$\pm0.7$}       &\gc  \textbf{30.1}\small{$\pm1.2$}\\
            & 200   &\rc 55.4\small{$\pm0.7$}     &\rc 57.5\small{$\pm1.1$}      & 48.6\small{$\pm0.8$}     &46.2\small{$\pm0.9$}       &\gc \textbf{43.6}\small{$\pm0.6$}     &\gc \textbf{45.7}\small{$\pm0.5$}     &\gc \textbf{28.7}\small{$\pm0.4$}      &\rc 39.4\small{$\pm0.6$}          & 25.5\small{$\pm0.5$}          & 29.8\small{$\pm0.2$}\\
        \midrule
            & 50  & 51.3\small{$\pm0.9$}   & 54.2\small{$\pm1.1$}     & 46.3\small{$\pm0.8$}     & 44.1\small{$\pm1.2$}       & 40.3\small{$\pm1.2$}     & 41.8\small{$\pm1.4$}     & 27.1\small{$\pm0.6$}     & 36.5\small{$\pm1.1$}    & 23.0\small{$\pm1.2$}      & 28.1\small{$\pm1.3$}  \\
        2  & 100   
        & 55.1\small{$\pm0.6$}     & 57.4\small{$\pm0.3$}      &\gc \textbf{49.5}\small{$\pm0.6$}     &\rc  46.3\small{$\pm0.9$}       &43.0\small{$\pm0.6$}     &\rc  45.4\small{$\pm1.1$}     &
        28.3\small{$\pm0.2$}      &\gc \textbf{39.7}\small{$\pm0.8$}          & 25.6\small{$\pm0.7$}          & 29.6\small{$\pm1.0$}\\
            & 200   
            &\gc \textbf{55.6}\small{$\pm0.9$}      &\gc \textbf{57.9}\small{$\pm0.5$}     & 49.4\small{$\pm0.3$}     & 46.0\small{$\pm0.1$}     &\rc  43.5\small{$\pm0.4$}     & 45.1\small{$\pm0.7$}     &\rc 28.6\small{$\pm0.2$}    & 39.3\small{$\pm0.8$}           &\gc  \textbf{25.9}\small{$\pm1.1$}          &\rc 29.9\small{$\pm0.6$}\\
        \midrule
            & 50  
            & 51.8\small{$\pm0.7$}       
            & 52.9\small{$\pm1.2$}     
            & 46.1\small{$\pm1.5$}    
            & 42.3\small{$\pm0.5$}     
            & 39.8\small{$\pm0.5$}     
            & 40.9\small{$\pm1.3$}     
            & 24.7\small{$\pm1.1$}    & 35.9\small{$\pm0.5$}        & 21.2\small{$\pm1.7$}        & 25.3\small{$\pm1.1$}\\
        4  & 100   
        & 53.3\small{$\pm0.8$}      
        & 54.2\small{$\pm1.1$}     
        & 47.3\small{$\pm1.2$}     & 41.8\small{$\pm1.7$}    
        & 42.7\small{$\pm0.6$}     & 27.7\small{$\pm0.5$}     
        & 27.1\small{$\pm1.0$}    & 27.0\small{$\pm0.9$}         & 22.5\small{$\pm1.4$}          & 26.4\small{$\pm1.2$}\\
            & 200   &\rc 55.0\small{$\pm1.0$}     &\rc 57.0\small{$\pm1.3$}     &\rc 48.1\small{$\pm1.6$}     &\rc  45.2\small{$\pm0.5$}     &\rc  42.1\small{$\pm1.4$}     &\rc  45.0\small{$\pm0.5$}     &\rc  27.2\small{$\pm0.9$}    &\rc 38.8\small{$\pm1.1$}         &\rc 24.6\small{$\pm0.5$}          &\rc 28.4\small{$\pm0.8$}\\
        \bottomrule
    \end{tabular}}
    \caption{Abltion study on layers combination and optimization allocation using TESLA. "Layers" refers to the number of layers per optimization space, "Optimization" refers to the number of SGD steps allocated in each optimization space.}
    \label{layers-TESLA}
\end{table*}

\begin{table*}[!htbp]
    \centering
    \renewcommand{\arraystretch}{1.2}
    \resizebox{\textwidth}{!}{
    \begin{tabular}{cccccccccccc}
        \toprule
        Layers & Optimization & ImNet-A & ImNet-B & ImNet-C & ImNet-D & ImNet-E & ImNette & ImWoof & ImNet-Birds & ImNet-Fruits & ImNet-Cats\\
        \midrule
            & 50  & 45.2\small{$\pm1.2$}     & 48.3\small{$\pm1.3$}     & 42.0\small{$\pm0.4$}     &36.2\small{$\pm0.7$}      & 35.0\small{$\pm0.8$}     & 35.8\small{$\pm1.1$}     & 22.7\small{$\pm1.0$}    & 33.5\small{$\pm0.5$}       & 21.1\small{$\pm1.5$}       & 22.7\small{$\pm0.8$}\\
        1 & 100   & 46.2\small{$\pm0.7$}     &\gc  \textbf{51.1}\small{$\pm0.4$}     & 43.3\small{$\pm1,1$}     & 37.2\small{$\pm0.5$}     & 36.6\small{$\pm0.9$}     & 36.7\small{$\pm1.3$}     & 22.9\small{$\pm0.8$}    &\rc 35.6\small{$\pm1.1$}       & 22.1\small{$\pm1.5$}       &\rc  23.8\small{$\pm0.7$}\\
            & 200   &\rc 46.5\small{$\pm0.9$}     & 50.7\small{$\pm1.1$}      &\rc  43.8\small{$\pm0.2$}     &\rc  37.3\small{$\pm0.7$}       &\gc \textbf{37.6}\small{$\pm0.7$}     &\rc 36.9\small{$\pm1.3$}     &\gc \textbf{24.3}\small{$\pm0.5$}      & 34.9\small{$\pm0.3$}          &\gc  \textbf{22.6}\small{$\pm1.3$}          & 23.6\small{$\pm0.7$}\\
        \midrule
            & 50  & 44.8\small{$\pm0.4$}   & 48.9\small{$\pm0.9$}     & 42.1\small{$\pm1.1$}     & 35.6\small{$\pm1.0$}       & 36.6\small{$\pm0.6$}     & 34.2\small{$\pm1.1$}     & 22.1\small{$\pm0.6$}     & 33.3\small{$\pm1.6$}    & 20.0\small{$\pm1.3$}      & 22.7\small{$\pm0.8$}  \\
        2  & 100   &\gc \textbf{46.9}\small{$\pm0.8$}     & 50.7\small{$\pm0.9$}     &\gc \textbf{43.9}\small{$\pm0.7$}       &\gc  \textbf{37.4}\small{$\pm0.4$}     & 37.2\small{$\pm0.3$}     & 36.9\small{$\pm0.8$}     &\rc  24.0\small{$\pm0.8$}    & 35.3\small{$\pm1.0$}         &\rc 22.4\small{$\pm1.1$}          & 24.1\small{$\pm0.9$}\\
            & 200   
            & 46.8\small{$\pm0.5$}      &\rc  50.8\small{$\pm0.3$}     & 43.4\small{$\pm0.6$}     & 37.0\small{$\pm1.3$}     &\rc 37.3\small{$\pm0.5$}     &\gc  \textbf{37.1}\small{$\pm0.7$}     & 23.8\small{$\pm1.3$}    &\gc  \textbf{35.6}\small{$\pm1.1$}           & 22.1\small{$\pm1.2$}          &\gc  \textbf{24.6}\small{$\pm1.3$}\\
        \midrule
            & 50  & 43.6\small{$\pm0.7$}       & 47.8\small{$\pm0.7$}     & 40.4\small{$\pm0.6$}     & 34.6\small{$\pm0.5$}     & 34.2\small{$\pm0.8$}     & 33.4\small{$\pm1.2$}     & 21.3\small{$\pm0.9$}    & 32.7\small{$\pm1.4$}        & 19.9\small{$\pm0.5$}        & 21.6\small{$\pm0.6$}\\
        4  & 100   
        & 45.7\small{$\pm0.7$}       & 49.4\small{$\pm0.9$}     
        & 43.1\small{$\pm1.1$}     & 36.1\small{$\pm1.3$}     
        & 36.4\small{$\pm0.8$}     & 35.2\small{$\pm0.6$}     
        &\rc  23.4\small{$\pm1.1$}    &\rc 34.7\small{$\pm0.5$}         & 21.3\small{$\pm1.1$}          & 23.5\small{$\pm1.3$}\\
            & 200   &\rc 46.3\small{$\pm0.8$}     &\rc  50.1\small{$\pm0.9$}     &\rc  43.2\small{$\pm0.7$}     &\rc  37.0\small{$\pm0.4$}     &\rc 36.8\small{$\pm1,6$}     &\rc 36.2\small{$\pm1.0$}     & 23.3\small{$\pm1.3$}    & 34.4\small{$\pm1.4$}         &\rc 21.6\small{$\pm0.8$}          &\rc 23.7\small{$\pm0.5$}\\
        \bottomrule
    \end{tabular}}
    \caption{Abltion study on layers combination and optimization allocation using DSA.}
    \label{layers-DSA}
\end{table*}

\subsection{Qualitative Interpretation using CAM}

We additionally introduce CAM \cite{muhammad2020eigen} to visualize the heatmap of class-relevant information in the synthetic images as shown in Figure \ref{cam change}, which also demonstrates our perspective from another aspect. The blue areas represent regions of class-relevant information, which can produce the largest gradient during the training process. Conversely, the red areas indicate regions of class-irrelevant information, with deeper colors signifying higher degrees of corresponding information. In the early stage of optimization, the class-relevant information of the main subject in the synthetic image produced by various distillation methods is compressed.  

Interestingly, for the gradient matching methods TESLA and DSA, which rely on long-range and short-range gradient matching respectively, the class-relevant information of the main subject remains unchanged when optimization space changes to later layers, while the gradient that can be produced by the image background (e.g., corners) are further decreased, as indicated by the deeper red color, even though the changes in the background are hardly observable by the naked eye during the optimization process. However, for the feature matching method DM, compared to the visualized kaleidoscope-like pattern, the visualization of corresponding CAM shows an unbalanced distribution and focuses on areas not typically observed by humans. We believe this phenomenon also explains the poorer performance of DM compared with gradient matching methods. Compared to the synthetic images with a centralized concentration of class-relevant information produced by TESLA and DSA, the images generated by DM are too diverse due to fitting all the features of the entire dataset including the class-irrelevant features, which is disadvantageous for training neural networks on tiny distilled datasets.

\subsection{Layers Combination and Optimization Allocation}

As discussed, we adopt a uniform sampling method that evaluates the synthetic dataset per $100$ optimization epochs (even less when using DM) to align with the evaluation method of the baseline (i.e., GLaD). Additionally, we decompose StylGAN-XL into $G_{11}\circ\cdots\circ G_{1}\circ G_{0}(\,\cdot\,)$ to align with the time complexity of the baseline. We present an ablation study on the allocation of optimization epochs per optimization space. Building on this, we further explore the impact of combining different numbers of intermediate layers into a single optimization space and allocating different numbers of optimization epochs to each optimization space on the performance of the synthetic dataset. For all distillation methods, we explore the impact of varying optimization spaces by using combinations of $1$, $2$,  and $4$ intermediate layers within each optimization space. Under the same optimization space setting, for TESLA and DSA, we investigated the effects of different numbers of optimization epochs allocated to each optimization space by using $50$, $100$, and $200$. For DM, due to the overfitting issue caused by feature matching, we used  $10$, $20$, and $50$ as the number of optimization epochs per optimization space.

The results for TESLA and DSA are shown in Table \ref{layers-TESLA} and Table \ref{layers-DSA}. Combining $1$ or $2$ intermediate layers as a single optimization space does not produce a significant impact on the performance, indicating that existing redundant feature spaces provided by GAN contribute little to the distillation tasks and may even lead to a negative effect. Under this setting, allocating $50$ optimization epochs per optimization space produces a clear phenomenon of optimization not converging. However, when the number of optimization epochs comes to $100$ or $200$, the optimization converges without significant performance differences. Achieved by implicitly selecting the optimal synthetic dataset through the proposed class-relevant feature distance metric, allowing us to avoid overfitting issues to some extent through a certain level of optimization path withdrawal. Therefore, we choose $100$ epochs as the basic setting to reduce time complexity in the actual training process. When using $4$ intermediate layers as an optimization space, the performance is decreased even when setting optimization epochs to $200$, indicating that too few feature domains could not provide sufficiently rich guiding information, forcing the optimization process to require more epochs to converge, demonstrating the superiority of our proposed H-PD in utilizing multiple feature domains.

\begin{table*}[!htbp]
    \centering
    \renewcommand{\arraystretch}{1.2}
    \resizebox{\textwidth}{!}{
    \begin{tabular}{cccccccccccc}
        \toprule
        Layers & Optimization & ImNet-A & ImNet-B & ImNet-C & ImNet-D & ImNet-E & ImNette & ImWoof & ImNet-Birds & ImNet-Fruits & ImNet-Cats\\
        \midrule
            & 10  &\rc 42.1\small{$\pm2.2$}     & 44.1\small{$\pm1.6$}     &\gc  \textbf{41.7}\small{$\pm1.7$}     &33.9\small{$\pm1.2$}      &\rc  31.3\small{$\pm1.9$}     &\rc  34.2\small{$\pm2.1$}     &\gc  \textbf{24.1}\small{$\pm1.4$}    &\gc  \textbf{29.7}\small{$\pm0.7$}       &\rc  24.1\small{$\pm1.6$}       & 22.6\small{$\pm1.3$}\\
        1 & 20   & 41.6\small{$\pm1.6$}     &\gc  \textbf{44.8}\small{$\pm1.8$}     
        & 41.3\small{$\pm1.4$}     &\rc  34.1\small{$\pm2.1$}     
        & 31.2\small{$\pm0.5$}     & 33.7\small{$\pm0.6$}     
        & 24.0\small{$\pm1.3$}    & 29.6\small{$\pm1.7$}       & 23.4\small{$\pm0.8$}       &\rc  23.7\small{$\pm1.9$}\\
            & 50   & 40.2\small{$\pm1.6$}     & 43.4\small{$\pm1.7$}      & 40.2\small{$\pm2.0$}     & 33.1\small{$\pm1.3$}       & 29.7\small{$\pm1.8$}     & 32.6\small{$\pm1.9$}     & 23.1\small{$\pm2.1$}      & 28.2\small{$\pm1.6$}          & 22.1\small{$\pm0.8$}          & 21.0\small{$\pm0.5$}\\
        \midrule
            & 10  & 41.4\small{$\pm1.7$}   & 43.5\small{$\pm1.3$}     & 40.4\small{$\pm0.9$}     & 34.1\small{$\pm1.3$}       & 31.3\small{$\pm1.8$}     & 33.6\small{$\pm1.7$}     & 22.4\small{$\pm1.6$}     & 28.3\small{$\pm2.1$}    & 23.1\small{$\pm1.7$}      & 22.9\small{$\pm1.5$}  \\
        2  & 20   &\gc  \textbf{42.8}\small{$\pm1.2$}     &\rc  44.7\small{$\pm1.3$}     
        &\rc  41.1\small{$\pm1.3$}       &\gc  \textbf{34.8}\small{$\pm1.5$}    
        &\gc  \textbf{31.9}\small{$\pm0.9$}     &\gc  \textbf{34.8}\small{$\pm1.0$}     
        &\rc  23.9\small{$\pm1.9$}    &\rc  29.5\small{$\pm1.5$}         &\gc  \textbf{24.4}\small{$\pm2.1$}          &\gc  \textbf{24.2}\small{$\pm1.1$}\\
            & 50   & 40.1\small{$\pm1.8$}      & 42.6\small{$\pm2.0$}     &  40.2\small{$\pm1.6$}     & 32.6\small{$\pm1.7$}     & 29.7\small{$\pm1.3$}     & 33.1\small{$\pm0.6$}     & 21.6\small{$\pm0.7$}    & 27.7\small{$\pm1.6$}           &22.2\small{$\pm1.3$}          &22.4\small{$\pm1.9$}\\
        \midrule
            & 10  & 39.9\small{$\pm1.4$}       & 42.5\small{$\pm1.0$}     &\rc  40.4\small{$\pm1.8$}     &\rc  32.4\small{$\pm1.6$}     & 30.1\small{$\pm2.4$}     & 32.7\small{$\pm2.3$}     & 20.9\small{$\pm1.6$}    & 27.5\small{$\pm2.2$}        &\rc  22.5\small{$\pm1.7$}        & 21.8\small{$\pm1.2$}\\
        4  & 20   &\rc  40.6\small{$\pm1.3$}       & 42.5\small{$\pm1.6$}     
        & 39.6\small{$\pm2.1$}     & 32.2\small{$\pm1.5$}     
        & 30.1\small{$\pm1.3$}     &\rc  32.9\small{$\pm1.8$}     
        & 21.6\small{$\pm1.5$}    & 27.3\small{$\pm1.2$}         & 21.7\small{$\pm2.3$}          & 22.3\small{$\pm1.6$}\\
            & 50   & 40.4\small{$\pm1.7$}     &\rc  42.7\small{$\pm1.3$}     & 39.9\small{$\pm1.2$}     & 32.0\small{$\pm1.4$}     &\rc  30.3\small{$\pm1.9$}     & 32.6\small{$\pm1.6$}     &\rc  22.0\small{$\pm1.1$}    &\rc  27.8\small{$\pm0.9$}         & 21.1\small{$\pm1.7$}          &\rc 22.6\small{$\pm1.4$}\\
        \bottomrule
    \end{tabular}}
    \caption{Abltion study on layers combination and optimization allocation using DM.}
    \label{layers-DM}
\end{table*}

The results for DM are shown in Table \ref{layers-DM}. Similar to TESLA and DSA, Combining $1$ or $2$ intermediate layers as a single optimization space results in similar performance, while combining $4$ intermediate layers as optimization space leads to a significant performance drop. However, under the same optimization space settings, an excessive number of optimization epochs often leads to a severe decline in performance when using DM as the distillation method. As aforementioned, DM is unable to focus on class-relevant information, which causes an irreversible loss of the main subject information in the synthetic image after deploying a large number of optimization epochs in a specific feature domain, which in turn leads to a situation where the informative guidance provided by subsequent feature domains could not be effectively incorporated into the synthetic image, resulting in performance degradation. In this case, even the proposed class-relevant feature distance could not effectively select a superior synthetic dataset. To align with the approach of decomposing GAN used in TESLA and DSA, we ultimately combine $2$ intermediate layers as an optimization space and deploy 20 optimization epochs as the experimental setting for DM.

\begin{table*}[!htbp]
    \centering
    \renewcommand{\arraystretch}{1.2}
    \resizebox{\textwidth}{!}{
    \begin{tabular}{cccccccccccc}
        \toprule
        Alg. & Searching Basis & ImNet-A & ImNet-B & ImNet-C & ImNet-D & ImNet-E & ImNette & ImWoof & ImNet-Birds & ImNet-Fruits & ImNet-Cats\\
        \midrule
            & -  &\rc 54.7\small{$\pm0.8$}     & 56.2\small{$\pm0.7$}     & 48.1\small{$\pm0.9$}     &\rc 45.4\small{$\pm0.9$}      & \rc 41.8\small{$\pm0.6$}     & 43.8\small{$\pm0.8$}     &\rc  28.1\small{$\pm1.0$}    &\rc  38.5\small{$\pm1.2$}       & 24.1\small{$\pm0.5$}       &\rc  28.7\small{$\pm0.9$}\\
        TESLA & Loss Value   & 53.6\small{$\pm0.9$}     & \rc 56.9\small{$\pm0.7$}     &\rc  48.3\small{$\pm0.8$}     &  45.0\small{$\pm0.6$}     & 41.0\small{$\pm1.2$}     &\rc  44.5\small{$\pm0.8$}     & 27.5\small{$\pm1.4$}    & 37.8\small{$\pm0.7$}       &\rc  25.1\small{$\pm0.9$}       & 27.6\small{$\pm1.0$}\\
            & \gc Feature Distance   &\gc \textbf{55.1}\small{$\pm0.6$}     &\gc \textbf{57.4}\small{$\pm0.3$}      &\gc \textbf{49.5}\small{$\pm0.6$}     &\gc \textbf{46.3}\small{$\pm0.9$}       &\gc \textbf{43.0}\small{$\pm0.6$}     &\gc \textbf{45.4}\small{$\pm1.1$}     &\gc \textbf{28.3}\small{$\pm0.2$}      &\gc \textbf{39.7}\small{$\pm0.8$}          &\gc \textbf{25.6}\small{$\pm0.7$}          &\gc \textbf{29.6}\small{$\pm1.0$}\\
        \midrule
            & -  & 45.9\small{$\pm0.7$}   &\rc  50.1\small{$\pm1.1$}     & 43.1\small{$\pm1.4$}     &\rc  36.9\small{$\pm0.8$}       &\rc  36.8\small{$\pm0.6$}     & 36.0\small{$\pm0.9$}     &\rc  23.6\small{$\pm0.8$}     &\rc 34.5\small{$\pm0.4$}    &\rc  21.9\small{$\pm0.8$}      &\rc 23.2\small{$\pm0.9$}  \\
        DSA  & Loss Value  &\rc 46.6\small{$\pm1.3$}     & 48.9\small{$\pm1.7$}     &\rc  43.6\small{$\pm1.1$}       & 36.1\small{$\pm1.2$}     & 36.6\small{$\pm0.5$}     &\rc 36.2\small{$\pm0.9$}     & 23.1\small{$\pm0.6$}    & 33.6\small{$\pm0.7$}         & 21.3\small{$\pm1.1$}          & 22.8\small{$\pm1.0$}\\
            &\gc Feature Distance   &\gc \textbf{46.9}\small{$\pm0.8$}      &\gc \textbf{50.7}\small{$\pm0.9$}     & \gc \textbf{43.9}\small{$\pm0.7$}     & \gc \textbf{37.4}\small{$\pm0.4$}     & \gc \textbf{37.2}\small{$\pm0.3$}     & \gc
            \textbf{36.9}\small{$\pm0.8$}     &\gc \textbf{24.0}\small{$\pm0.8$}    &\gc \textbf{35.3}\small{$\pm1.0$}           &\gc \textbf{22.4}\small{$\pm1.1$}          &\gc \textbf{24.1}\small{$\pm0.9$}\\
        \midrule
            & -  &\rc 42.4\small{$\pm1.6$}       & 44.2\small{$\pm2.1$}     &\rc  41.0\small{$\pm1.2$}     & 34.0\small{$\pm1.2$}     &\rc  31.1\small{$\pm1.0$}     & 34.5\small{$\pm2.1$}     & 23.1\small{$\pm0.9$}    &\rc 29.0\small{$\pm1.5$}        & 24.1\small{$\pm1.4$}        &\rc  22.6\small{$\pm1.5$}\\
        DM  & Loss Value  & 41.6\small{$\pm1.8$}       & \rc 44.4\small{$\pm1.4$}     & 40.7\small{$\pm2.1$}     & \rc 34.6\small{$\pm1.7$}     & 30.1\small{$\pm1.3$}     & \rc 34.5\small{$\pm1.3$}     &\rc  23.6\small{$\pm1.2$}    & 28.7\small{$\pm1.3$}         &\rc  24.4\small{$\pm1.3$}          & 21.2\small{$\pm1.2$}\\
            &\gc Feature Distance  &\gc \textbf{42.8}\small{$\pm1.2$}     &\gc \textbf{44.7}\small{$\pm1.3$}     &\gc \textbf{41.1}\small{$\pm1.3$}     & \gc\textbf{34.8}\small{$\pm1.5$}     &\gc \textbf{31.9}\small{$\pm0.9$}     &\gc \textbf{34.8}\small{$\pm1.0$}     &\gc \textbf{23.9}\small{$\pm1.9$}    &\gc \textbf{29.5}\small{$\pm1.5$}         &\gc \textbf{24.4}\small{$\pm2.1$}          &\gc \textbf{24.2}\small{$\pm1.1$}\\
        \bottomrule
    \end{tabular}}
    \caption{Quantitative results on searching basis. "-" refers to not employing a searching strategy, "Loss Value" refers to directly using corresponding loss function value as the searching basis, "Feature Distance" refers to the use of proposed class-relevant distance as a searching basis}
    \label{performance:loss-distance}
\vspace{-1.em}
\end{table*}

\subsection{Ablation Study on Searching Strategy}

\begin{figure}[!htbp]
\centering
\resizebox{0.8\linewidth}{!}{
\includegraphics{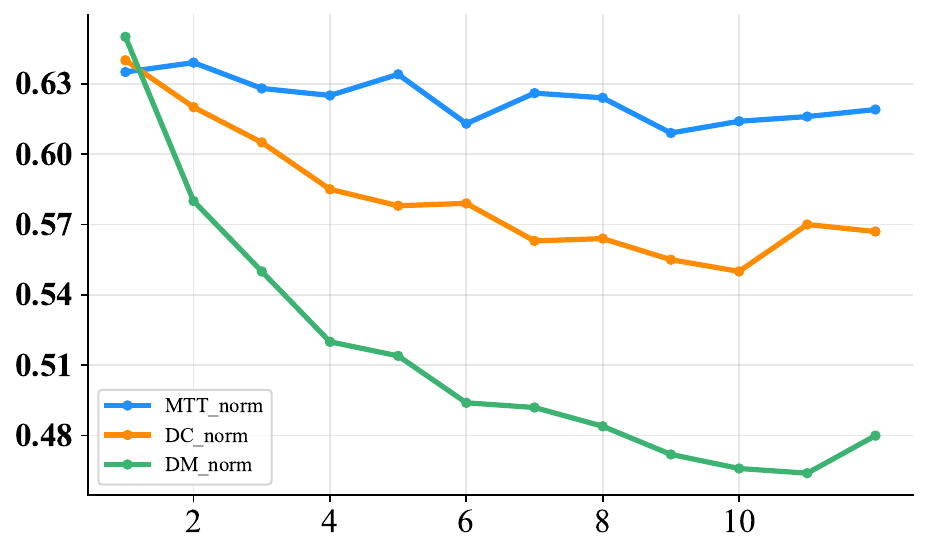}}
    \captionof{figure}{Quantitative results of loss function value using different distillation methods. Note that we normalize all the values for clear comparison.}
    \label{vis-loss}
\end{figure}

To better utilize the informative guidance provided by multiple feature domains, we propose class-relevant feature distance as an evaluation metric for implicitly selecting the optimal synthetic dataset. We demonstrate the ablation study using different implicit evaluation metrics, as shown in Table \ref{performance:loss-distance}, the metric we proposed outperforms the use of loss function value corresponding to the distillation methods as the metric under all settings. It is worth noting that, although the accuracy of the model trained on the synthetic dataset can be used as an explicit evaluation metric for the data distillation task, the evaluation process incurred much greater time overhead than the distillation task itself, rendering it impractical for actual training processes.

To explore the principle of the superiority of class-relevant feature distance, we first discussed the respective limitations of directly using existing distillation loss function value as the evaluation metric. The tendency of different distillation loss functions is shown in Figure \ref{vis-loss}. For TESLA, the loss function is obtained by calculating the distance between the student network parameters and the teacher network parameters. However, in order to consider diversity, TESLA selects a random initialization method when initializing the student network parameters, and the expert trajectory also comes from the training process of models with different initialization, leading to a significant fluctuation caused by utilizing different initialization parameters. For DSA, the loss function utilizes neural network gradients as guidance. However, when IPC=1, the proxy neural network used in each optimization process is randomly initialized, causing DSA to face the same issue as TESLA, where the loss function is affected by network parameter initialization. As for DM, the loss function is obtained from the feature distance between the dataset features extracted by randomly initialized networks, resulting in the same impact of network initialization parameters on this loss function. Additionally, DM suffers from severe overfitting in the later stages of optimization due to fitting to the useless features. In summary, the loss functions corresponding to the three distillation methods could not serve as effective evaluation metrics due to the excessive diversity.

\begin{figure}[!htbp]
\centering
\resizebox{0.8\linewidth}{!}{
\includegraphics{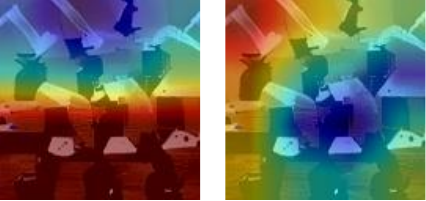}}
    \captionof{figure}{The visualization comparison of CAM between pre-trained model and random model using DM.}
    \label{cam comparison}
\end{figure}

Distinguished from existing distillation methods, where the loss function is influenced by the need to fit diversity, our proposed class-relevant feature distance effectively addresses this issue by using CAM, which is calculated by utilizing a pre-trained neural network, and we utilize a ResNet-18 trained on ImageNet-1k as a proxy model for computing CAM. As shown in Figure \ref{cam comparison}, we demonstrate the difference between the visualization obtained using the pre-trained model and those obtained using a randomly initialized model. The observation indicates that there is a significant difference in the regions of interest for the two, by utilizing a pre-trained model with fixed parameters, we can better identify the feature regions that are beneficial for the classification task (i.e., larger gradients). Therefore, our proposed metric successfully leverages this strong supervisory signal to achieve data selection while eliminating the strong correlation between the loss function and the proxy model parameters.

\begin{table}[!htbp]
    \centering
    \resizebox{\linewidth}{!}{%
    \begin{tabular}{lccccc}
        \toprule
        Method  & ImNet-A  & ImNet-B & ImNet-C & ImNet-D & ImNet-E \\
        \midrule
         GLaD-TESLA  & 50.7\small{$\pm0.4$} & 51.9\small{$\pm1.3$} & 44.9\small{$\pm0.4$} & 39.9\small{$\pm1.7$} & 37.6\small{$\pm0.7$}\\
         + Average Initialization  & \gc \textbf{51.9}\small{$\pm1.0$} &\gc  \textbf{53.5}\small{$\pm0.7$} &\gc   \textbf{46.1}\small{$\pm0.9$} &\gc   \textbf{41.0}\small{$\pm0.7$} &\gc  \textbf{39.1}\small{$\pm1.0$}\\
        \midrule
        GLaD-DSA  & 44.1\small{$\pm2.4$} &\gc  \textbf{49.2}\small{$\pm1.1$} &\gc  \textbf{42.0}\small{$\pm0.6$} & 35.6\small{$\pm0.9$} &\gc  \textbf{35.8}\small{$\pm0.9$}\\
         + Average Initialization  &\gc   \textbf{45.4}\small{$\pm0.6$} & 48.9\small{$\pm0.8$} &  40.6\small{$\pm0.7$} &\gc \textbf{36.4}\small{$\pm0.5$} & 34.8\small{$\pm0.3$}\\
        \midrule
        GLaD-DM  & 41.0\small{$\pm1.5$} & 42.9\small{$\pm1.9$} & 39.4\small{$\pm1.7$} &\gc \textbf{33.2}\small{$\pm1.4$} & 30.3\small{$\pm1.3$}\\
         + Average Initialization  &\gc   \textbf{41.5}\small{$\pm1.2$} &\gc   \textbf{43.2}\small{$\pm1.6$} &\gc  \textbf{39.9}\small{$\pm1.7$} &  32.2\small{$\pm0.9$} &\gc  \textbf{30.8}\small{$\pm1.3$}\\
        \bottomrule
    \end{tabular}}
    \caption{Ablation study of average noise initialization on GLaD.} 
    \label{glad noise}
\end{table}

\subsection{Ablation Study on Average Noise Initialization}

To investigate the effect of using averaged noise as initialization, we conduct ablation experiments on both GLaD and H-PD respectively. As shown in Table \ref{glad noise}, averaged noise often provides a significant gain for GLaD. Indicating that using averaged noise as input tends to produce images with reduced bias that conform to the statistical characteristics of the real dataset, implying that images generated from averaged noise are usually centered within the real dataset. As aforementioned, since GLaD neglects the informative guidance from the earlier layers, leading to a lack of optimization for the main subject of the synthetic image, averaged noise can to some extent replace this operation.

\begin{table}[!htbp]
    \centering
    \resizebox{\linewidth}{!}{%
    \begin{tabular}{lccccc}
        \toprule
        Method  & ImNet-A  & ImNet-B & ImNet-C & ImNet-D & ImNet-E \\
        \midrule
         H-PD-TESLA  & 54.1\small{$\pm0.5$} & 56.8\small{$\pm0.4$} & 48.9\small{$\pm1.3$} & 45.0\small{$\pm0.7$} & 42.1\small{$\pm0.6$}\\
         + Average Initialization  &\gc  \textbf{55.1}\small{$\pm0.6$} &\gc  \textbf{57.4}\small{$\pm0.3$} &\gc   \textbf{49.5}\small{$\pm0.6$} &\gc   \textbf{46.3}\small{$\pm0.9$} &\gc  \textbf{43.0}\small{$\pm0.6$}\\
        \midrule
        H-PD-DSA  & 46.5\small{$\pm1.0$} & 50.4\small{$\pm0.4$} &\gc  \textbf{44.5}\small{$\pm0.6$} &\gc  \textbf{37.7}\small{$\pm1.1$} & 36.9\small{$\pm0.7$}\\
         + Average Initialization  &\gc  \textbf{46.9}\small{$\pm0.8$} &\gc  \textbf{50.7}\small{$\pm0.9$} &  43.9\small{$\pm0.7$} &  37.4\small{$\pm0.4$} &\gc  \textbf{37.2}\small{$\pm0.3$}\\
        \midrule
        H-PD-DM  & 42.6\small{$\pm1.6$} & 44.5\small{$\pm0.9$} &\gc  \textbf{42.3}\small{$\pm1.4$} & 34.5\small{$\pm1.1$} & \gc \textbf{32.3}\small{$\pm1.3$}\\
         + Average Initialization  &\gc  \textbf{42.8}\small{$\pm1.2$} &\gc   \textbf{44.7}\small{$\pm1.3$} & 41.1\small{$\pm1.3$} &\gc   \textbf{34.8}\small{$\pm1.5$} &  31.9\small{$\pm0.9$}\\
        \bottomrule
    \end{tabular}}
    \caption{Ablation study of average noise initialization on  H-PD.} 
    \label{hglad noise}
\end{table}

As shown in Table \ref{hglad noise}, average noise initialization provides only a limited improvement for H-PD on TESLA, while using DSA and DM, averaged noise is closer to random initialization. The observation aligns with our perspective that H-PD requires optimization through all layers of the GAN, which has already led to optimization for the main subject information that conforms to the constraints of the loss function during the early stages of training. The role of averaged noise is then reduced to merely providing samples that better conform to statistical characteristics, which is also why we still employ averaged noise for H-PD to obtain a training-free optimization starting point.  

Additionally, since DSA tends to optimize towards classification boundary samples or noisy samples, and DM tends to substantially modify synthetic datasets to achieve feature maximum mean discrepancy optimization, neither GLaD nor H-PD with average noise initialization can effectively improve the performance on DSA and DM. Nevertheless, TESLA is most effective in preserving the primary subject information in the synthetic images, which allows for the averaging of noise and the achievement of a relatively stable improvement.

\section{Experimental Details} 
\label{implementation details}

\begin{table*}[!htbp]
    \centering
    \renewcommand{\arraystretch}{1.2}
    \resizebox{\textwidth}{!}{
    \begin{tabular}{ccccccccccc}
        \toprule
        Dataset & 0 & 1 & 2 & 3 & 4 & 5 & 6 & 7 & 8 & 9\\
        \midrule
        ImNet-A & Leonberg & \makecell{Probiscis\\Monkey} & Rapeseed & \makecell{Three-Toed\\Sloth} & \makecell{Cliff\\Dwelling} & \makecell{Yellow\\Lady’s Slipper} & Hamster & Gondola & Orca & Limpkin \\
        \midrule
        ImNet-B & Spoonbill & Website & Lorikeet & Hyena & Earthstar & Trollybus & Echidna & Pomeranian & Odometer & \makecell{Ruddy\\Turnstone}\\
        \midrule
        ImNet-C & \makecell{Freight\\Car} & Hummingbird & Fireboat & \makecell{Disk\\Brak} & \makecell{Bee\\Eater} & \makecell{Rock\\Beauty} & Lion & \makecell{European\\Gallinule} & \makecell{Cabbage\\Butterfly} & Goldfinch\\
        \midrule
        ImNet-D & Ostrich & Samoyed & Snowbird & \makecell{Brabancon\\Griffon} & Chickadee & Sorrel & Admiral & \makecell{Great\\Gray Owl} & Hornbill & Ringlet\\
        \midrule
        ImNet-E & Spindle & Toucan & \makecell{Black\\Swan} & \makecell{King\\Penguin} & \makecell{Potter’s\\Wheel} & Photocopier & Screw & Tarantula & Oscilloscope & Lycaenid\\
        \midrule
        ImNette & Tench & \makecell{English\\Springer} & \makecell{Cassette\\Player} & Chainsaw & Church & \makecell{}French Horn & \makecell{Garbage\\Truck} & \makecell{Gas\\Pump} & \makecell{Golf\\Ball} & Parachute\\
        \midrule
        ImWoof & \makecell{Australian\\Terrier} & \makecell{Border\\Terrier} & Samoyed & Beagle & Shih-Tzu & \makecell{English\\Foxhound} & \makecell{Rhodesian\\Ridgeback} & Dingo & \makecell{Golden\\Retriever} & \makecell{English\\Sheepdog}\\
        \midrule
        ImNet-Birds & Peacock & Flamingo & Macaw & Pelican & \makecell{King\\Penguin} & \makecell{Bald\\Eagle} & Toucan & Ostrich & \makecell{Black\\Swan} & Cockatoo\\
        \midrule
        ImNet-Fruits & Pineapple & Banana & Strawberry & Orange & Lemon & Pomegranate & Fig & Bell Pepper & Cucumber & \makecell{Granny\\Smith Apple}\\
        \midrule
        ImNet-Cats & \makecell{Tabby\\Cat} & \makecell{Bengal\\Cat} & \makecell{Persian\\Cat} & \makecell{Siamese\\Cat} & \makecell{Egyptian\\Cat} & Lion & Tiger & Jaguar & \makecell{Snow\\Leopard} & Lynx\\
        \bottomrule
    \end{tabular}}
    \caption{Corresponding class names in each ImageNet-Subsets. The visualizations follow the same order.}
    \label{class name}
\end{table*}

\begin{table*}[!tbp]
    \centering
    \resizebox{\linewidth}{!}{%
    \begin{tabular}{ccccccccccc}
            \toprule 
            Dataset & IPC & \makecell{Synthetic\\steps} & \makecell{Expert\\epochs} & \makecell{Max expert\\epoch} & \makecell{Trajectory\\number} & \makecell{Learning rate\\\small(Learning rate)} & \makecell{Learning rate\\\small(Teacher)} & \makecell{Learning rate\\\small(Latent w)} & \makecell{Learning rate\\\small(Latent f)} & \makecell{Steps\\per space} \\
            \midrule
            \multirow{2}{*}{CIFAR-10} & \large$1$   & \large$20$ & \large$3$ & \large$50$ & \large$100$ & \large$10^{-6}$ & \large$10^{-2}$ & \large$10^{1}$ & \large$10^{4}$ & \large$100$ \\
                                      \cmidrule{2-11}
                                      & \large$10$   & \large$20$ & \large$3$ & \large$50$ & \large$100$ & \large$10^{-6}$ & \large$10^{-2}$ & \large$10^{1}$ & \large$10^{4}$ & \large$100$ \\
            \midrule
            ImageNet-Subset & \large$1$   & \large$20$ & \large$3$ & \large$15$ & \large$200$ & \large$10^{-6}$ & \large$10^{-2}$ & \large$10^{1}$ & \large$10^{4}$ & \large$100$ \\
            \bottomrule
          \end{tabular}}
    \caption{TESLA hyper-parameters} 
    \label{TESLA hyper-parameters}
\end{table*}

\begin{table}[!tbp]
    \centering
    \resizebox{\linewidth}{!}{%
    \begin{tabular}{ccccc}
            \toprule
            Dataset & IPC & \makecell{Learning rate\\\small(Latent w)} & \makecell{Learning rate\\\small(Latent f)} & \makecell{Steps\\per space} \\
            \midrule
            \multirow{2}{*}{CIFAR-10} & \large$1$ & \large$10^{-2}$ & \large$10^{1}$ & \large$20$ \\
                                      \cmidrule{2-5}
                                      & \large$10$ & \large$10^{-2}$ & \large$10^{1}$ & \large$20$ \\
            \midrule
            \multirow{2}{*}{ImageNet-Subset} & \large$1$ & \large$10^{-2}$ & \large$10^{1}$ & \large$20$ \\
                                      \cmidrule{2-5}
                                      & \large$10$ & \large$10^{-2}$ & \large$10^{1}$ & \large$20$ \\
            \bottomrule
        \end{tabular}}
    \caption{DM hyper-parameters} 
    \label{DM hyper-parameters}
    \vspace{-1.0em}
\end{table}

\begin{table}[!htbp]
    \centering
    \resizebox{\linewidth}{!}{%
    \begin{tabular}{ccccccc}
            \toprule
            Dataset & IPC & \makecell{inner\\loop} & \makecell{outer\\loop}& \makecell{Learning rate\\\small(Latent w)} & \makecell{Learning rate\\\small(Latent f)} & \makecell{Steps\\per space} \\
            \midrule
            \multirow{2}{*}{CIFAR-10} & \large$1$ & \large$1$ & \large$1$ &\large$10^{-3}$ & \large$10^{0}$ & \large$100$ \\
                                      \cmidrule{2-7}
                                      & \large$10$ & \large$50$ & \large$10$ &\large$10^{-3}$ & \large$10^{0}$ & \large$100$ \\
            \midrule
            \multirow{2}{*}{ImageNet-Subset} & \large$1$ & \large$1$ & \large$1$ & \large$10^{-3}$ & \large$10^{0}$ & \large$100$ \\
                                      \cmidrule{2-7}
                                      & \large$10$ & \large$50$ & \large$10$ & \large$10^{-3}$ & \large$10^{0}$ & \large$100$ \\
            \bottomrule
        \end{tabular}}
    \caption{DSA hyper-parameters} 
    \label{DSA hyper-parameters}
\end{table}

\subsection{Dataset}
\label{dataset settings}
We evaluate H-PD on various datasets, including a low-resolution dataset CIFAR10\cite{krizhevsky2009learning} and a large number of high-resolution datasets ImageNet-Subset.
\begin{itemize}
  \item CIFAR-10 consists of $32\times32$ RGB images with 50,000 images for training and 10,000 images for testing. It has 10 classes in total and each class contains 5,000 images for training and 1,000 images for testing.
  \item ImageNet-Subset is a small dataset that is divided out from the ImageNet\cite{deng2009imagenet} based on certain characteristics. By aligning with the previous work, we use the same types of subsets: ImageNette (various objects)\cite{howard2019smaller}, ImageWoof (dogs)\cite{howard2019smaller}, ImageFruit (fruits) \cite{cazenavette2022dataset}, ImageMeow (cats) \cite{cazenavette2022dataset}, ImageSquawk (birds) \cite{cazenavette2022dataset},  and ImageNet-[A, B, C, D, E] (based on ResNet50 performance) \cite{cazenavette2023generalizing}. Each subset has 10 classes. The specific class name in each Imagenet-Subset is shown in Table \ref{class name}.
\end{itemize}

\subsection{Network Architecture}
For the comparison of same-architecture performance, we employ a convolutional neural network ConvNet-3 as the backbone network as well as the test network. For low-resolution datasets, we employ a 3-depth convolutional neural network ConvNet-3 as the backbone network, consisting of three basic blocks and one fully connected layer. Each block includes a $3\times3$ convolutional layer, instance normalization~\cite{ulyanov2016instance}, ReLU non-linear activation, and a $2\times2$ average pooling layer with a stride of 2. After the convolution blocks, a linear classifier outputs the logits. For high-resolution datasets, we employ a 5-depth convolutional neural network ConvNet-5 as the backbone network for $128\times128$ resolution,  ConvNet-5 has five duplicate blocks, which is as the same as that in ConvNet-3. For $256\times256$ resolution, we employ ConvNet-6 as the backbone network.

For the comparison of cross-architecture performance, we also follow the previous work: ResNet-18 \cite{he2016deep}, VGG-11 \cite{simonyan2014very}, AlexNet \cite{krizhevsky2012imagenet}, and ViT \cite{dosovitskiy2020image} from the DC-BENCH~\cite{cui2022dc} resource.

\subsection{Implementation details}

The implementation of our proposed H-PD is based on the open-source code for GLaD~\cite{cazenavette2023generalizing}, which is conducted on NVIDIA GeForce RTX 3090.

To ensure fairness, we utilize identical hyperparameters and optimization settings as GLaD. In our experiments, we also adopt the same suite of differentiable augmentations (originally from the DSA codebase~\cite{zhao2021dataset}), including color, crop, cutout, flip, scale, and rotate. We use an SGD optimizer with momentum, $\ell_{2}$ decay. The entire distillation process continues for 1200 epochs. We evaluate the performance of the synthetic dataset by training 5 randomly initialized networks on it. 

To obtain the expert trajectories used in MTT, we train a backbone model from scratch on the real dataset for 15 epochs of SGD with a learning rate of $10^{-2}$, a batch size of 256, and no momentum or regularization. To maintain the integration of different distillation methods, we do not use the ZCA whitening on both high-resolution datasets and low-resolution datasets different from previous work\cite{cazenavette2022dataset}, which leads to a same-architecture performance drop, please note that our proposed H-PD still outperforms under the same setting. Different from GLaD which records 1000 expert trajectories for the MTT method, we only record 200 expert trajectories and thus largely reduce the computational costs. Additionally, while GLaD performs 5k optimization epochs on the synthetic dataset using MTT as the distillation method, we only perform 1k optimization epochs and achieve better performance both on same-architecture and cross-architecture settings, further proving the superiority of our H-PD. The detailed hyperparameters are shown in Table \ref{DM hyper-parameters}, Table \ref{DSA hyper-parameters} and Table \ref{TESLA hyper-parameters}.

\section{More Visualizations}
\label{more visualization}
\begin{figure*}[!ht]
\centering
\resizebox{\textwidth}{!}{
\includegraphics{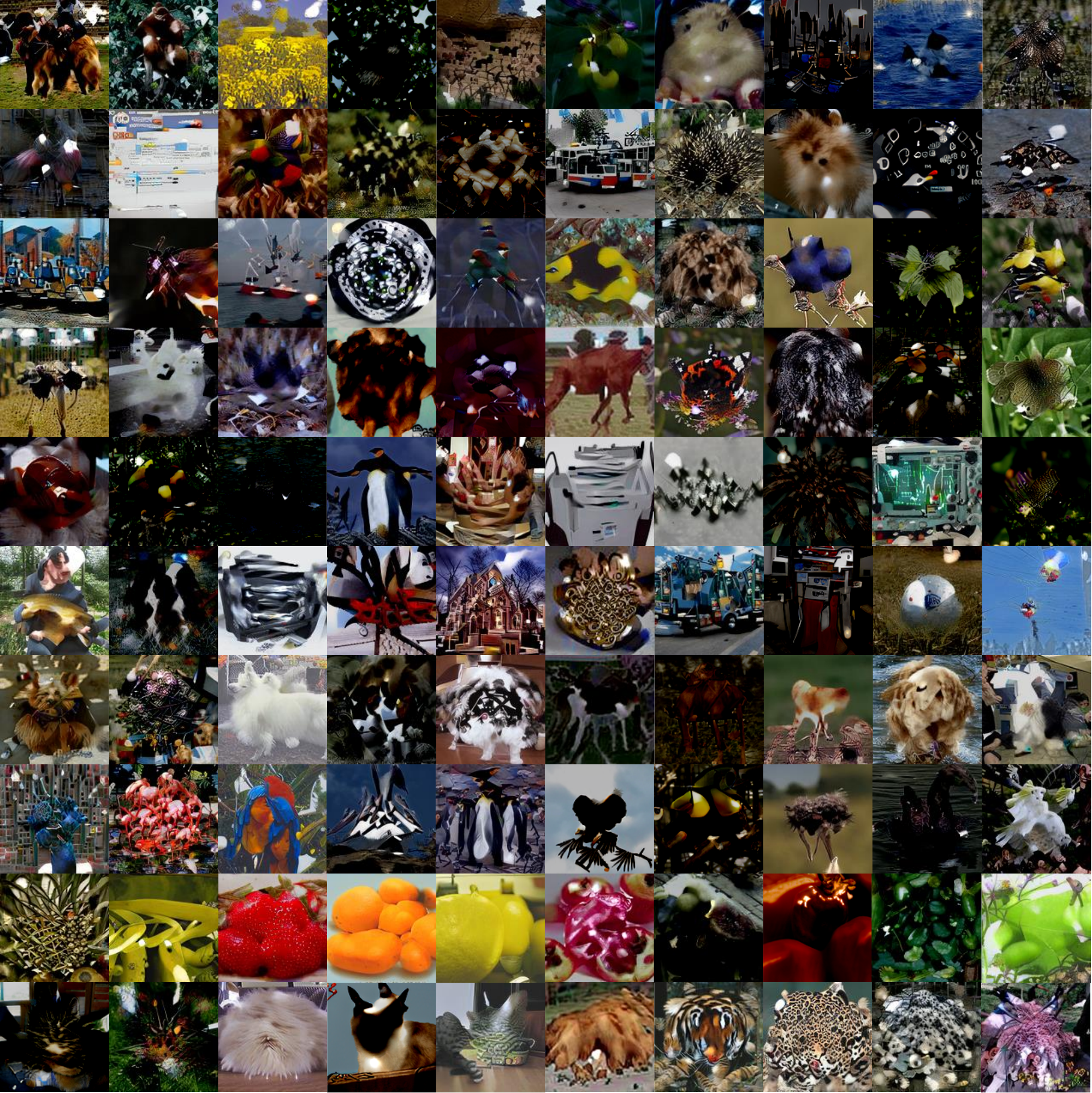}}
\caption{More visualization of the synthetic datasets using TESLA.}
\label{visualization-TESLA}
\end{figure*}

\begin{figure*}[!ht]
\centering
\resizebox{\textwidth}{!}{
\includegraphics{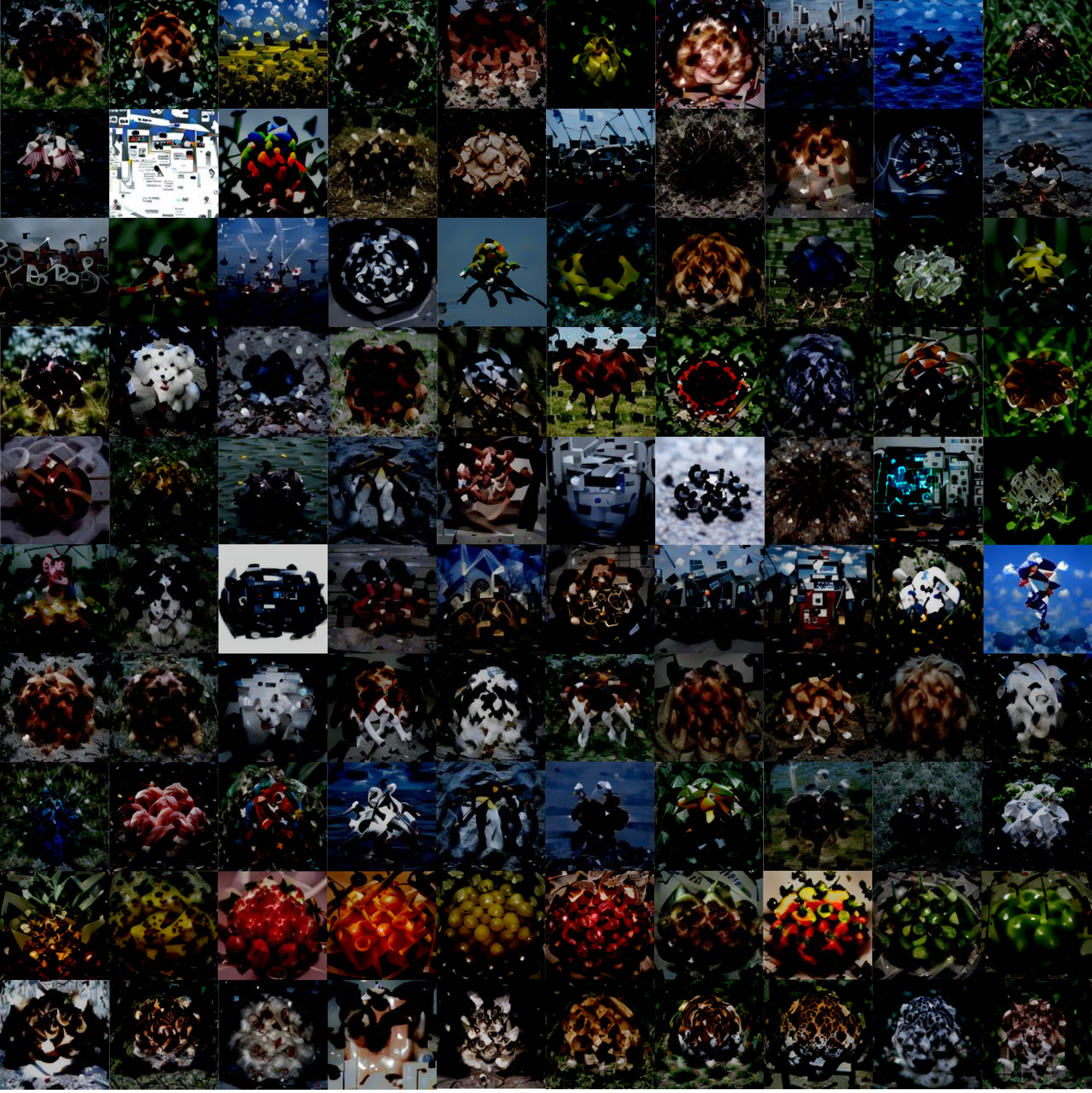}}
\caption{More visualization of the synthetic datasets using DSA.}
\label{visualization-DSA}
\end{figure*}

\begin{figure*}[!ht]
\centering
\resizebox{\textwidth}{!}{
\includegraphics{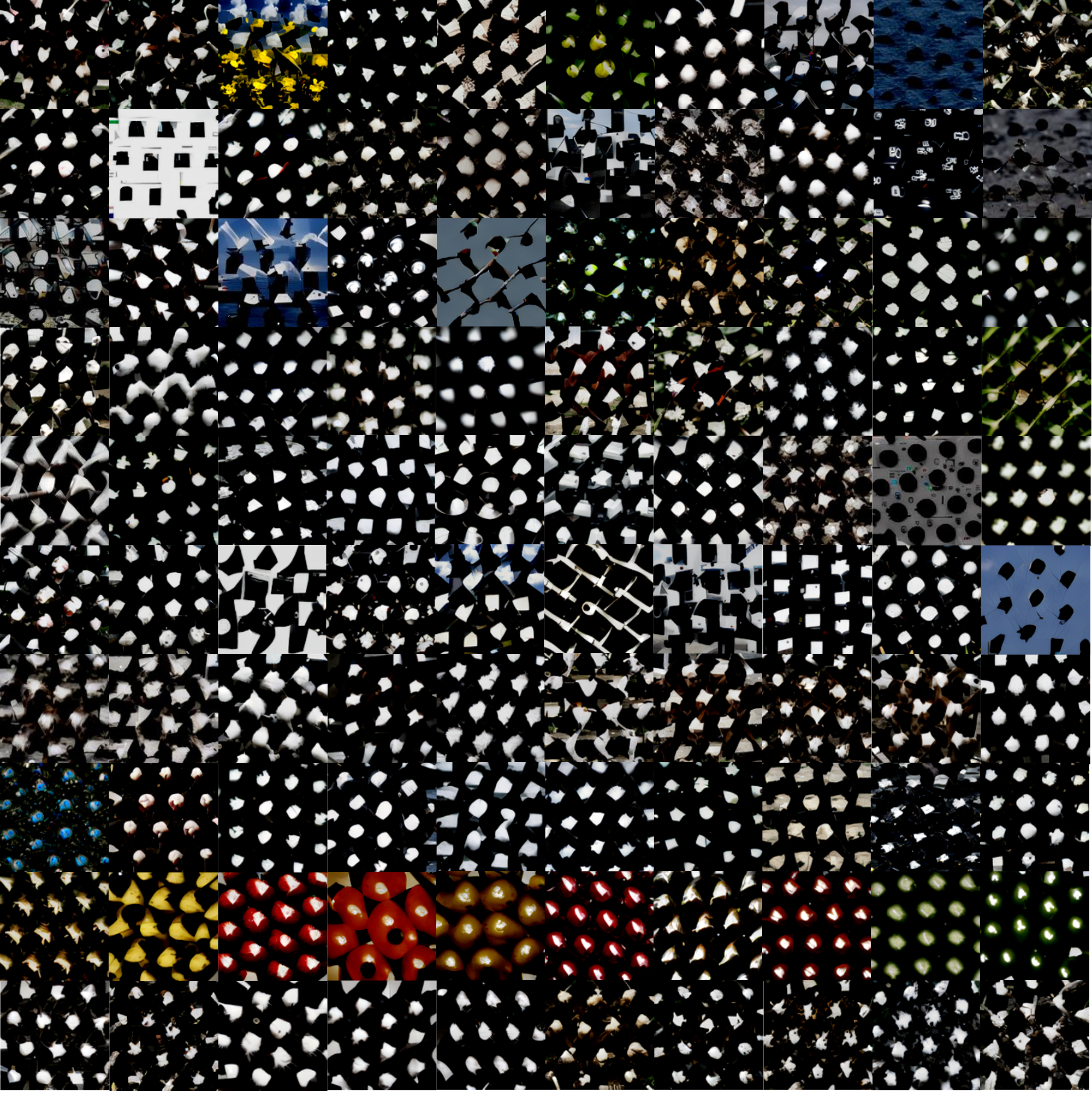}}
\caption{More visualization of the synthetic datasets using DM.}
\label{visualization-DM}
\end{figure*}

We provide additional visualizations of synthetic datasets generated by H-PD using diverse distillation methods, as shown in Figure \ref{visualization-TESLA}, Figure \ref{visualization-DSA}, and Figure \ref{visualization-DM}.

\clearpage

\clearpage
\clearpage


\end{document}